\documentclass[preprint,12pt]{elsarticle}

\usepackage[utf8]{inputenc} 
\usepackage{hyperref}   
\usepackage{url}            
\usepackage{booktabs}       
\usepackage{amsfonts}       
\usepackage{nicefrac}       
\usepackage{microtype}      
\usepackage[table,xcdraw]{xcolor}

\usepackage{multirow}
\usepackage[T1]{fontenc} 
\usepackage{amsmath}
\usepackage[cmintegrals]{newtxmath}
\usepackage{bm} 
\usepackage{algpseudocode}
\usepackage{algorithm}
\usepackage{varwidth}
\algnewcommand{\LineComment}[1]{\State \(\triangleright\) #1}
\usepackage{caption}
\usepackage{subcaption}
\usepackage{enumitem}
\setlist{nolistsep}
\newcolumntype{L}[1]{>{\raggedright\arraybackslash}m{#1}}
\newcolumntype{C}[1]{>{\centering\arraybackslash}m{#1}}
\newcolumntype{R}[1]{>{\raggedleft\arraybackslash}m{#1}}

\usepackage{lineno}

\journal{Neural Networks}

\begin{document}

\begin{frontmatter}



\title{Tree-CNN: A Hierarchical Deep Convolutional Neural Network for Incremental Learning}


\author{Deboleena Roy\corref{cor1}}
\ead{roy77@purdue.edu}
\author{Priyadarshini Panda}
\ead{pandap@purdue.edu}
\author{Kaushik Roy}
\ead{kaushik@purdue.edu}

\cortext[cor1]{Corresponding author}
\address{Department of Electrical and Computer Engineering, Purdue University, West Lafayette, IN 47907, USA}
\begin{abstract}
Over the past decade, Deep Convolutional Neural Networks (DCNNs) have shown remarkable performance in most computer vision tasks. These tasks traditionally use a fixed dataset, and the model, once trained, is deployed as is. Adding new information to such a model presents a challenge due to complex training issues, such as ``catastrophic forgetting'', and sensitivity to hyper-parameter tuning. However, in this modern world, data is constantly evolving, and our deep learning models are required to adapt to these changes. In this paper, we propose an adaptive hierarchical network structure composed of DCNNs that can grow and learn as new data becomes available. The network grows in a tree-like fashion to accommodate new classes of data, while preserving the ability to distinguish the previously trained classes. The network organizes the incrementally available data into feature-driven super-classes and improves upon existing hierarchical CNN models by adding the capability of self-growth. The proposed hierarchical model, when compared against fine-tuning a deep network, achieves significant reduction of training effort, while maintaining competitive accuracy on CIFAR-10 and CIFAR-100. 
\end{abstract}


\end{frontmatter}


\section{Introduction}
\label{introduction}
In recent years Deep Convolutional Neural Networks (DCNNs) have emerged as the leading architecture for large scale image classification \cite{Rawat2017}. In 2012, AlexNet \cite{krizhevsky2012imagenet}, an 8 layer Deep CNN,  won the ImageNet Large Scale Visual Recognition Challenge (ISLVRC) and catapulted DCNNs into the spotlight. Since then, they have dominated ISLVRC and have performed extremely well on popular image datasets such as MNIST \cite{lecun1998gradient,wan2013regularization}, CIFAR-10/100 \cite{krizhevsky2009learning}, and ImageNet \cite{russakovsky2015imagenet}.

Today, with increased access to large amounts of labeled data (eg. ImageNet \cite{russakovsky2015imagenet} contains 1.2 million images with 1000 categories), supervised learning has become the leading paradigm in training DCNNs for image recognition. Traditionally, a DCNN is trained on a dataset containing a large number of labeled images. The network learns to extract relevant features and classify these images. This trained model is then used on real world unlabeled images to classify them. In such training, all the training data is presented to the network during the same training process. However, in real world, we hardly have all the information at once, and data is, instead, gathered incrementally over time. This creates the need for models that can learn new information as it becomes available. In this work, we try to address the challenge of learning on such incrementally available data in the domain of image recognition using deep networks.

A DCNN embeds feature extraction and classification in one coherent architecture within the same model. Modifying one part of the parameter space immediately affects the model globally \cite{Xiao2014}. Another problem of incrementally training a DCNN is the issue of ``catastrophic forgetting'' \cite{Goodfellow2013}. When a trained DCNN is retrained exclusively over new data, it results in the destruction of existing features learned from earlier data. This mandates using previous data when retraining on new data.

To avoid catastrophic forgetting, and to leverage the features learned in previous task, this work proposes a network made of CNNs that grows hierarchically as new classes are introduced. The network adds the new classes like new leaves to the hierarchical structure. The branching is based on the similarity of features between new and old classes. The initial nodes of the \textit{Tree-CNN} assign the input into coarse super-classes, and as we approach the leaves of the network, finer classification is done. Such a model allows us to leverage the convolution layers learned previously to be used in the new bigger network. 

The rest of the paper is organized as follows. The related work on incremental learning in deep neural networks is discussed in Section \ref{sec:relatedwork}. In Section \ref{sec:method} we present our proposed network architecture and incremental learning method. In Section \ref{sec:experiments}, the two experiments using CIFAR-10 and CIFAR-100 datasets are described. It is followed by a detailed analysis of the performance of the network and its comparison with transfer learning and fine tuning in Section \ref{sec:results}. Finally, Section \ref{sec:discussion} discusses the merits and limitations of our network, our findings, and possible opportunities for future work. 

\section{Related Work}
\label{sec:relatedwork}

The modern world of digitized data produces new information every second \cite{john2014big}, thus fueling the need for systems that can learn as new data arrives. Traditional deep neural networks are static in that respect, and several new approaches to incremental learning are currently being explored. ``One-shot learning'' \cite{fei2006one} is a Bayesian transfer learning technique, that uses very few training samples to learn new classes.  Fast R-CNN \cite{Girshick_2015}, a popular framework for object detection, also suffers from ``catastrophic forgetting''. One way to mitigate this issue is to use a frozen copy of the original network compute and balance the loss when new classes are introduced in the network \cite{Shmelkov_2017}. ``Learning without Forgetting'' \cite{li2017learning} is another method that uses only new task data to train the network while preserving the original capabilities. The original network is trained on an extensive dataset, such as ImageNet \cite{russakovsky2015imagenet}, and the new task data is a much smaller dataset. ``Expert Gate'' \cite{aljundi2016expert} adds networks (or experts) trained on new tasks sequentially to the system and uses a set of gating autoencoders to select the right network (``expert'') for the given input.  Progressive Neural Networks \cite{rusu2016progressive} learn to solve complex sequences of task by leveraging prior knowledge with lateral connections. Another recent work on incremental learning in neural networks is ``iCaRL'' \cite{rebuffi2017icarl}, where they built an incremental classifier that can potentially learn incrementally over an indefinitely long time period. 

It has been observed that initial layers of a CNN learn very generic features \cite{Sarwar2017} that has been exploited for transfer learning \cite{yosinski2014transferable},, \cite{sarwar2017incremental}. Common features, that are shared between images, have been used previously to build hierarchical classifiers. These features can be grouped semantically, such as in \cite{panda2017semantic}, or be feature-driven, such as ``FALCON'' \cite{panda2017falcon}. Similar to the progression of complexity of convolutional layers in a DCNN, the upper nodes of a hierarchical CNN classify the images into coarse super-classes using basic features, like grouping green-colored objects together, or humans faces together. Then deeper nodes perform finer discrimination, such as ``boy'' v/s ``girl'' , ``apples'' v/s ``oranges'', etc.  Such hierarchical CNN models have been shown to perform at par or even better than standard DCNNs \cite{Yan_2015}. ``Discriminative Transfer Learning'' \cite{srivastava2013discriminative} is one of the earliest works where classes are categorized hierarchically to improve network performance. Deep Neural Decision Forests \cite{kontschieder2015deep} unified decision trees and deep CNN's to build a hierarchical classifier. ``HD-CNN'' \cite{Yan_2015}, is a hierarchical CNN model that is built by exploiting the common feature sharing aspect of images. However, in these works, the dataset is fixed from the beginning, and prior knowledge of all the classes and their properties is used to build a hierarchical model. 

In our work, \textit{Tree-CNN} starts out as a single root node and generates new hierarchies to accommodate the new classes. Images belonging to the older dataset are required during retraining, but by localizing the change to a small section of the whole network, our method tries to reduce the training effort and complexity. In \cite{Xiao2014}, a similar approach is applied, where the new classes are added to the old classes, and divided into two super-classes, by using an error-based model. The initial network is cloned to form two new networks which are fine tuned over the two new super-classes. While their motivation was a ``divide-and-conquer'' approach for large datasets, our work tries to incrementally grow with new data over multiple learning stages. In the next section, we lay out in detail our design principle, network topology and the algorithm used to grow the network.

\section{Incremental Learning Model}
\label{sec:method}
\subsection{Network Architecture}

\begin{figure}[h]
\centering
  \includegraphics[width=0.6\linewidth]{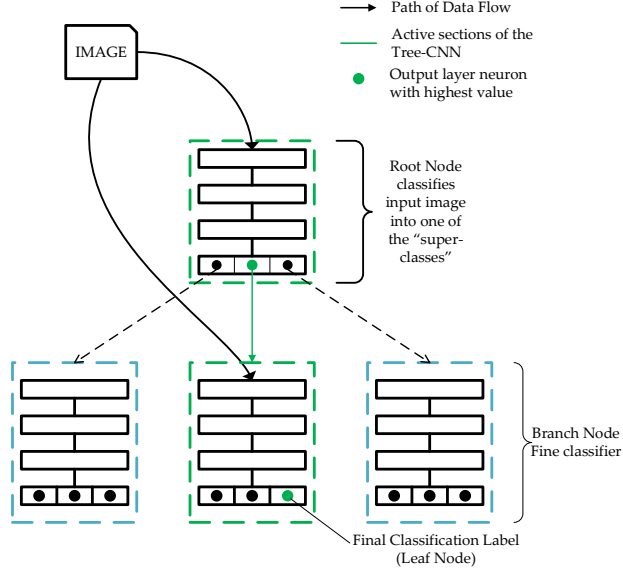}
  \caption{A generic model of 2-level Tree-CNN: The output of the root node is used to select the branch node at the next level.}
  \label{fig:generic_model}
\end{figure}

Inspired from hierarchical classifiers, our proposed model, \textit{Tree-CNN} is composed of multiple nodes connected in a tree-like manner. Each node (except leaf nodes) has a DCNN which is trained to classify the input to the node into one of it's children. The root node is the highest node of the tree, where the first classification happens. The image is then passed on to its child node, as per the classification label. This node further classifies the image, until we reach a leaf node, the last step of classification. Branch nodes are intermediary nodes, each having a parent and two or more children. The leaf node is the last level of the tree. Each leaf node is uniquely associated to a class and no two leaf nodes have the same class. Fig. \ref{fig:generic_model} shows the root node and branch nodes for a two-stage classification network. Each output of the second level branch node is a leaf node, which is the output node of the branch CNN. The inference methodology of such a network is given by Algorithm \ref{algo:inference-tree}.
\begin{algorithm}
\caption{Tree-CNN: At Inference}
\label{algo:inference-tree}
\begin{algorithmic}[1]
\State $I = $ Input Image, $node =$ Root Node of the Tree
\Procedure {ClassPredict}{$I, node$}
\State $count =$ \# of children of node
\If {$count = 0$} 
\State $label =$ class label of the node
\State return $label$
\Else 
\State $nextNode = EvaluateNode(I, node)$
\State $\blacktriangleright$ returns the address of the child node of highest output neuron
\State return $ClassPredict(I, nextNode)$
\EndIf
\EndProcedure
\end{algorithmic}
\end{algorithm}

\subsection{The Learning Algorithm}
\label{subsec:algorithm}

\begin{figure}[h]
\centering
  \includegraphics[width=1\linewidth]{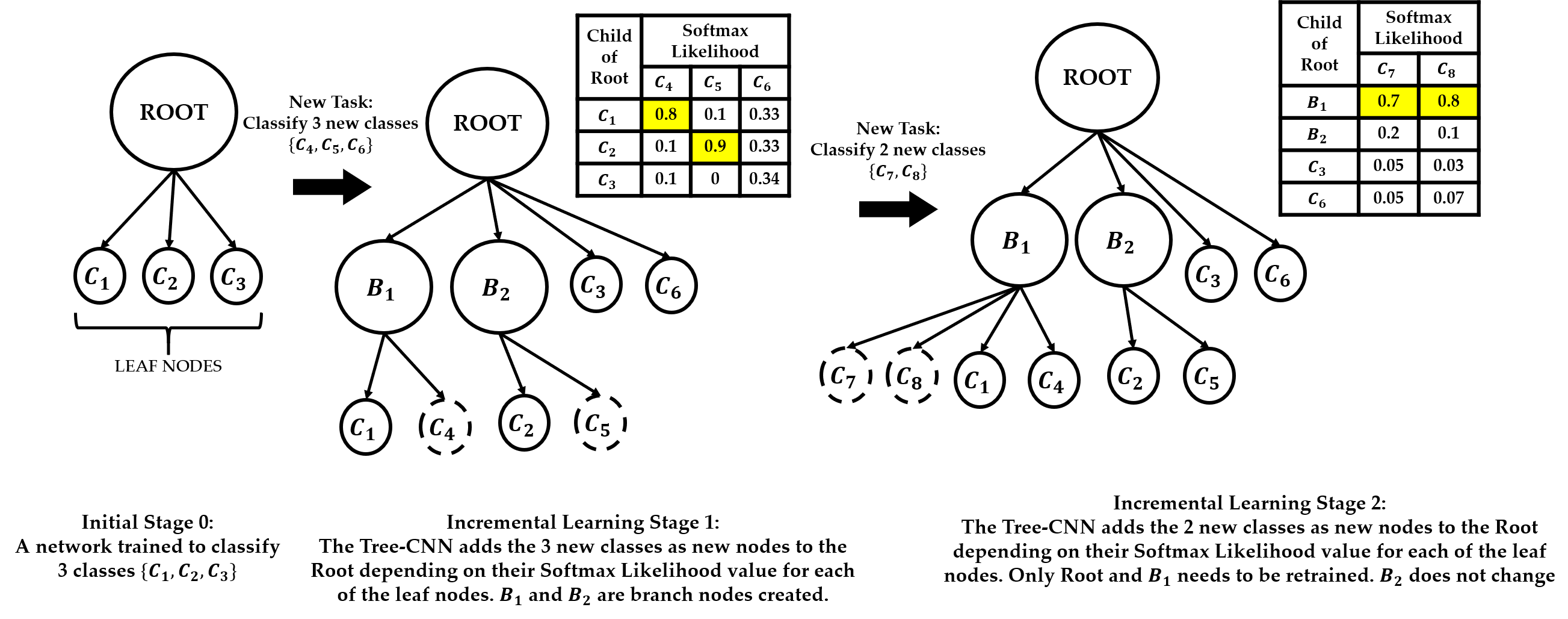}
  \caption{An example illustrating multiple incremental learning stages of the Tree-CNN. The network starts as a single root node, and expands as new classes are added.}
  \label{fig:incremental_learning}
\end{figure}

We start with the assumption that we have a model that is already trained to recognize a certain number of objects. The model could be hierarchical with multiple CNNs or could be just a single CNN acting as a root node with multiple leaf nodes. A new task is defined as learning to identify images belonging to $M$ new classes. We start at the root node of our given model, and we provide a small sample of images ($\sim10\%$) from the new training set as input to this node. 

We obtain a $3$ dimensional matrix from the output layer, $O^{K\times M\times I}$, where, $K$ is the number of children of the root node, $M$ is the number of new classes, and $I$ is the number of sample images per class. $O(k,m,i)$ denotes the output of the $k^{th}$ neuron for the $i^{th}$ image belonging to the $m^{th}$ class where $k\in[1,K]$, $m\in[1,M]$, and $i\in[1,I]$. $O_{avg}^{K\times M}$ is the average of the outputs over $I$ images. Softmax likelihood is computed over $O_{avg}$ (eq. \ref{eq:softmax1}) to obtain the likelihood matrix $L^{K\times M}$ (eq. \ref{eq:softmax2}). 
\begin{gather} 
\label{eq:softmax1}
O_{avg}(k,m) = \sum\limits_{i=1}^I \frac{O(k,m,i)}{I} \\
\label{eq:softmax2}
L(k,m) = \frac{e^{O_{avg}(k,m)}}{\sum\limits_{k=1}^K e^{O_{avg}(k,m)}} \\
\end{gather}

We generate an ordered list $S$ from $L^{K\times M}$, having the following properties
\begin{itemize}
\item The list $S$ has $M$ objects. Each object corresponds uniquely to one of the new $M$ classes.
\item Each object $S[i]$ has the following attributes:
\begin{itemize}
\item $S[i].label =$ label of the new class
\item $S[i].value = [v_1, v_2, v_3]$, top 3 average softmax ($o_{avg}$) output values for that class in descending order, $v_1\geq v_2 \geq v_3$
\item $S[i].nodes = [n_1, n_2, n_3]$, output nodes corresponding to the softmax outputs $v_1$, $v_2$, $v_3$
\end{itemize}
\item $S$ is ordered in the decreasing value of $S[i].value[1]$
\end{itemize}

The ordering is done to ensure that new classes with high likelihood values are added first to \textit{Tree-CNN}. Softmax likelihood is used instead of number of images that get classified as each of the child nodes because it translates the output layer's response to the images into an exponential scale and helps us better identify how similar an image is to one of the already existing labels. After constructing $S$, we look at its first element, $S[1]$, and take one of the 3 actions.
\begin{itemize}
\item[i.] \textbf{Add the new class to an existing child node:} If $v_{1}$ is greater than the next value ($v_{2}$) by a threshold, $\alpha$ (a design specification), that class indicates a strong resemblance/association with a particular child node. The new class is added to corresponding child node $n_{1}$.
\item[ii.] \textbf{Merge two child nodes to form a new child node and add the new class to this node:} If there are more than 1 child nodes that the new class has a strong likelihood for, we can combine them to form a new child node. It happens when $v_1 - v_2 < \alpha$, and $v_2 - v_3 > \beta$ (another threshold, defined by the user). For example, if the top 3 likelihood values were $v_1 = 0.48$, $v_2 = 0.45$, and $v_3 = 0.05$. Then, provided $n_2$ is a leaf node, we merge $n_2$ into $n_1$, and add the new class to $n_1$.
\item[iii.] \textbf{Add the new class as a new child node:} If the new class doesn't have a likelihood value that is greater than other values by a good margin ($v_1 - v_2 < \alpha, v_2 - v_3 < \beta$), or all child nodes are full, the network expands horizontally by adding the new class as a new child node. This node will be a leaf node. 
\end{itemize}
\vspace{10 pt}

As the root node keeps adding new branches and sub-branches, the branch nodes with more children tend to get heavier. Incoming new classes tend to have a higher softmax likelihood for branch nodes with greater number of children. To prevent the \textit{Tree-CNN} from becoming lop-sided, one can set the maximum number of children a branch node can have. 

When calculating $L(k,m)$, we substitute $e^{O_{avg}(k,m)}$ with 0 for those $k$ branches that are `full', i.e have reached the limit for number of children per branch. We assign $S[1].label$ a location in the \textit{Tree-CNN} depending on its $value$. After that, we remove the column corresponding to that class from $L(k,m)$, we check for ``full'' branch nodes, and modify $L(k,m)$ for those output nodes. Finally we generate the ordered list $S$, and again apply our conditions on the new $S[1]$ to determine where it is added to the root node. This is done iteratively till all new classes are assigned a location under the root node. 

The pseudo-code is outlined in Algorithm \ref{algo:grow-tree}. We also illustrate a toy example of incremental learning in Tree-CNN with Fig. \ref{fig:incremental_learning}. The network starts as a single CNN that can classify 3 classes, ${C_{1}, C_{2}, C_{3}}$. We want to increase the network capability by adding 3 new classes. In the first incremental learning stage, the softmax likelihood table $L$ is generated, as shown in the figure. $C_{4}$ and $C_{5}$ are added to the leaf nodes containing $C_{1}$ and $C_{2}$ respectively, converting them into branch nodes $B_{1}$ and $B_{2}$, as per condition (i). For $C_{6}$, the 3 likelihood values are $v_{1} = 0.34, v_{2} = 0.33, v_{3} = 0.33$. It satisfies neither condition (i) nor condition (ii), thus it is added as a new node to the root, as per condition (iii). Again, as new information is available, we want the Tree-CNN to be able to recognize 2 new image classes, $C_{7}$, and $C_{8}$. Both the new classes satisfy $v_{1} - v_{2} > \alpha (=0.1)$. Thus, both the classes are added to $B_{1}$. While this example is for a two level Tree-CNN, the algorithm can potentially be extended to deeper Tree-CNN models. 

\begin{algorithm}
\caption{Grow Tree-CNN}
\label{algo:grow-tree}
\begin{algorithmic}[1]
\State $L =$ Likelihood Matrix
\State $maxChildren$ = max. number of children per branch node
\State $RootNode =$ Root Node of the Tree-CNN
\Procedure {GrowTree}{L, Node}
\State $S = GenenerateS(L, Node, maxChildren)$
\While{$S$ is not Empty}
\State $\blacktriangleright$ Get attributes of the first object
\State $[label, value, node] = GetAttributes(S[1])$
\If{$value[1] - value[2] > \alpha$}
\State $\blacktriangleright$ The new class has a strong preference for $n_1$
\State $\blacktriangleright$ Adds $label$ to $node[1]$
\State $RootNode = AddClasstoNode(RootNode, label, node[1])$
\Else
\If{$value[2] - value[3] > \beta$}
\State $\blacktriangleright$ The new class has similar strong preference $n_1$ and $n_2$
\State $Merge = CheckforMerge(Node, node[1], node[2])$
\State $\blacktriangleright$ $Merge$ is $True$ only if $node[2]$ is a leaf node, and, 
\State $\blacktriangleright$ the \# of children of $node[1]$ less than $maxChildren - 1$ 
\If {$Merge$}
\State $\blacktriangleright$ Merge $node[2]$ into $node[1]$
\State $RootNode = MergeNode(RootNode, node[1], node[2])$
\State $RootNode = AddClasstoNode(RootNode, label, node[1])$
\Else
\State $\blacktriangleright$ Add new class to the smaller output node
\State $sNode=$ Node with lesser children ($node[1], node[2]$)
\State $RootNode = AddClasstoNode(RootNode, label, sNode)$
\EndIf
\Else 
\State $\blacktriangleright$ Add new class as a new Leaf node to Root Node
\State $RootNode = AddNewNode(RootNode, label)$
\EndIf
\EndIf
\State $\blacktriangleright$ Remove the columns of the added class from $L$
\State $\blacktriangleright$ Remove the rows of ``full'' nodes from $L$
\State $\blacktriangleright$ Regenerate $S$
\State $S = GenenerateS(L, Node, maxChildren)$
\EndWhile
\EndProcedure
\end{algorithmic}
\end{algorithm}

To create deeper Tree-CNN models, once the ``Grow-Tree'' algorithm is completed for the $M$ classes at the root node, one can move to the next level of the tree. The same process is applicable on the child nodes that now have new classes to be added to them. The decision on how to grow the tree is semi-supervised: the algorithm itself decides how to grow the tree, given the constraints by the user. We can limit parameters such as maximum children for a node, maximum depth for the tree, etc. as per our system requirements. 

Once the new classes are allotted locations in the tree, supervised gradient descent based training is performed on the modified/new nodes. This saves us from modifying the whole network, and only affected portions of the network require retraining/fine-tuning. At every incremental learning stage, the root node is trained on all the available data as it needs to learn to classify all the objects into the new branches. During inference, a branch node is activated only when the root node classifies the input to that branch node. If an incorrect classification happens at Root Node, for example it classifies an image of a car into the ``Animal Node'' (CIFAR-10 example, Sec 4.1), irrespective of what the branch node classifies it as, it would still be an incorrect classification. Hence we only train the branch node with the classes it has been assigned to. If there is no change in the branch node's look up table at an incremental learning stage, it is left as is.

\subsubsection*{Handling input labels inside the Tree-CNN} The dataset available to the user will have unique labels assigned to each of it's object classes. However, the root and branch nodes of the Tree-CNN tend to group/merge/split these classes as required by the algorithm. To ensure label consistency, each node of the Tree-CNN maintains it's own ``LabelsTransform'' lookup table. For example, when a new class is added to one of the pre-existing output nodes of a root node, the lookup table is updated with new class being assigned to that output node. Similarly when a new class is added as a new node, the class label and the new output node is added as a new entry to the lookup table. Every class is finally associated with a unique leaf node, hence leaf nodes do not require a look up table. Whenever two nodes are merged, the node with lower average softmax value (say, node A) gets integrated with the node with the higher average softmax value (say, node B) for the new class in consideration. If the two softmax values are equal, it is chosen at random. At the root node level, the lookup table is modified as follows: The class labels that were assigned to node A, will now be assigned to node B. The look up table of merged node B will add these class labels from node A as new entries and assign them to new leaf nodes. 

\section{The Experimental Setup}
\label{sec:experiments}
\subsection{Adding Multiple New Classes (CIFAR-10)}\label{subsec:cifar10}
\begin{figure}[h]
\centering
  \includegraphics[width=0.8\linewidth]{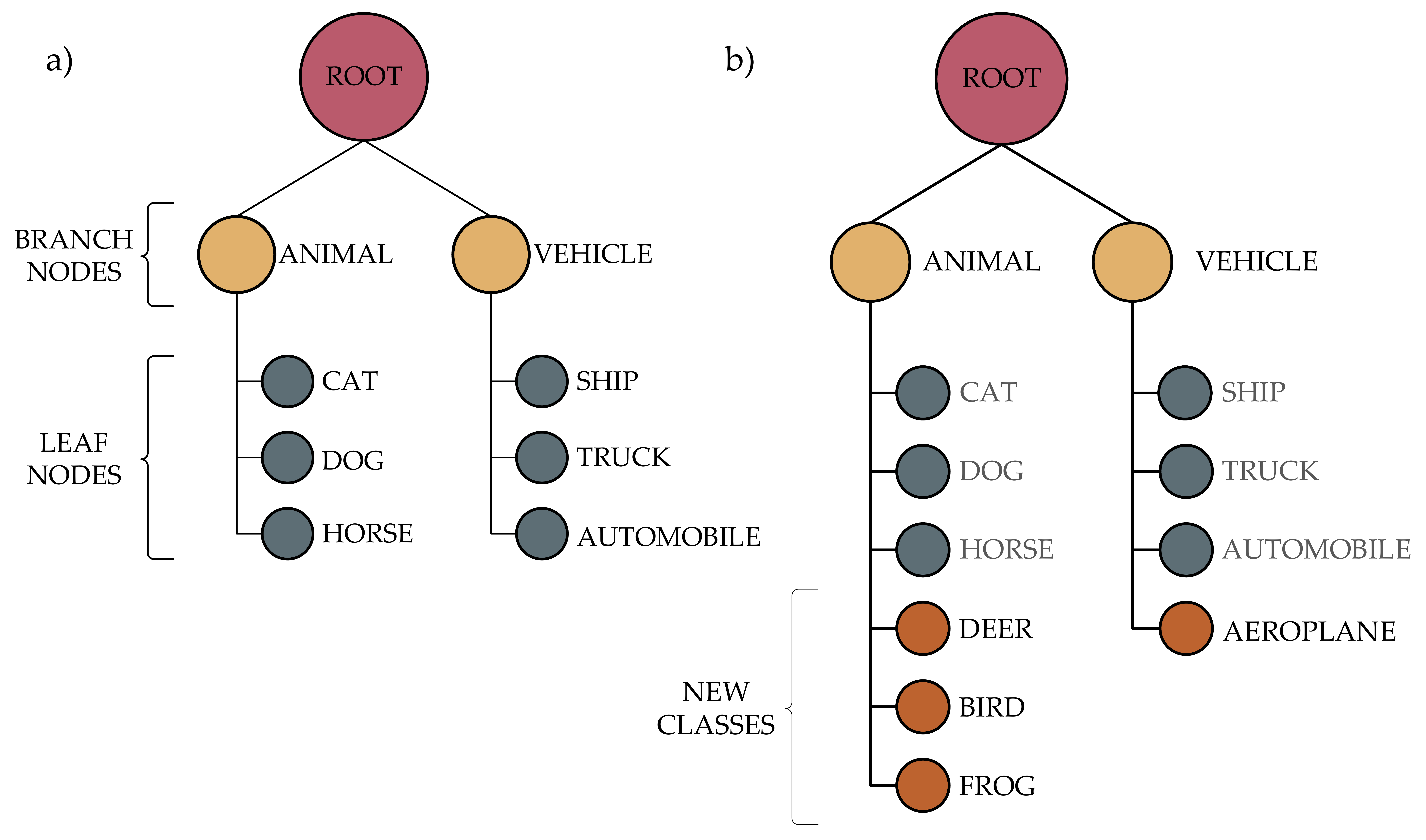}
  \caption{Graphical representation of Tree-CNN for CIFAR-10 \textbf{a)} before incremental learning, \textbf{b)} after incremental learning}
  \label{fig:cifar10treecompare}
\end{figure}

\begin{table}[h!]
\parbox[t]{0.3\linewidth}{
\centering
\caption{Root Node Tree-CNN (CIFAR-10)}
\label{table:cifar-10-root-node}
\scalebox{1}{
\begin{tabular}{|c|}
\hline
Input 32$\times$32$\times$3\\ \hline
CONV-1 \\
64 5$\times$5 ReLU \\ \hline
{[}2 2{]} Max Pooling\\ \hline
CONV-2 \\
128 3$\times$3 ReLU \\
Dropout 0.5
\\ 128 3$\times$3 ReLU \\ \hline
{[}2 2{]} Max Pooling\\ \hline
FC \\ 
8192$\times$512 ReLU\\ 
Dropout 0.5\\ 
512$\times$128 ReLU\\ 
Dropout 0.5\\
128$\times$2 ReLU \\ \hline
Softmax Layer \\ \hline
\end{tabular}
}}
\hfill
\parbox[t]{0.3\linewidth}{
\centering
\caption{Branch Node Tree-CNN (CIFAR-10)}
\label{table:cifar-10-branch-node}
\scalebox{1}{
\begin{tabular}{|c|}
\hline
Input 32$\times$32$\times$3\\ \hline
CONV-1\\
32 5$\times$5 ReLU\\ \hline
{[}2 2{]} Max Pooling \\ \hline
Dropout 0.25\\ \hline
CONV-2\\
64 5$\times$5 ReLU\\ \hline
{[}2 2{]} Max Pooling\\ \hline
Dropout 0.25\\ \hline
CONV-3\\
64 3$\times$3 ReLU\\ \hline
{[}2 2{]} Avg Pooling\\ \hline
Dropout 0.25\\ \hline
FC \\
1024$\times$128 ReLU\\ Dropout 0.5\\ 
128$\times$N ReLU\\ ($N=\#$ of Classes) \\ \hline
Softmax Layer \\ \hline
\end{tabular}
}}
\hfill
\parbox[t]{0.3\linewidth}{
\centering
\caption{Network B}
\label{table:Network_B}
\scalebox{1}{
\begin{tabular}{|c|}
\hline
Input 32$\times$32$\times$3\\ \hline
CONV-1\\
64 3$\times$3 ReLU\\
Dropout 0.5\\
64 3$\times$3 ReLU\\ \hline
{[}2 2{]} Max Pooling\\ \hline
CONV-2\\
128 3$\times$3 ReLU\\ 
Dropout 0.5\\
128 3$\times$3 ReLU\\ \hline
{[}2 2{]} Max Pooling\\ \hline
CONV-3\\
256 3$\times$3 ReLU\\
Dropout 0.5\\
256 3$\times$3 ReLU\\ \hline
{[}2 2{]} Max pooling\\ \hline
CONV-4\\
512 3$\times$3 ReLU\\
Dropout 0.5\\
512 3$\times$3 ReLU\\ \hline
{[}2 2{]} Avg Pooling\\ \hline
FC\\
2048$\times$1024 ReLU\\
Dropout 0.5\\
1024$\times$1024 ReLU\\
Dropout 0.5\\
1024$\times$N\\
(N$= \#$ of Classes)\\ \hline
\end{tabular}
}}
\end{table}

\subsubsection{Dataset}CIFAR-10 dataset \cite{krizhevsky2009learning}, having 10 mutually exclusive classes, was used for this experiment.  The network is first trained on 6 classes, and then learns the remaining 4 classes in the next learning stage.  

\subsubsection{The Network Initialization} 

The \textit{Tree-CNN} for CIFAR-10 starts out as a two level network with a root node with two branch nodes as shown in Fig \ref{fig:cifar10treecompare}. The six initial classes of CIFAR-10 are grouped into ``Vehicles'' and ``Animals'' and the CNN (Table \ref{table:cifar-10-root-node}) at the root is trained to classify input images into these two categories. Each of the two branch nodes has a CNN (Table \ref{table:cifar-10-branch-node}) that does finer classification into leaf nodes. Fig. \ref{fig:cifar10treecompare}a) represents the initial model of \textit{Tree-CNN A}. This experiment illustrates, that given provided a 2-level \textit{Tree-CNN}, how the learning model can add new classes. The root node achieves a testing accuracy of $98.73\%$, while the branch nodes, ``Animals'' and ``Vehicles'', achieve $86\%$ and $94.43\%$ testing accuracy respectively. Overall, the network achieves a testing accuracy of $89.10\%$.

\subsubsection{Incremental Learning} The remaining four classes are now introduced as the new learning task. $50$ images  per class ($10\%$ of the training set) are selected at random , and shown to the root node. We obtain the $L$ matrix, which is a $2\times4$ matrix with each element $l_{ij}\in(0,1)$. The $1^{st}$ row of the matrix indicates the softmax likelihood of each of the $4$ classes as being classified as ``Vehicles'', while the second row presents the same information for ``Animals''. In this experiment, $\alpha$ is set at 0 (Algorithm \ref{algo:grow-tree}), and the network is bound to take only one action: add the new class to one of the two child nodes. The branch node with higher likelihood value adds the new class to itself. The \textit{Tree-CNN} before and after addition of these 4 classes is shown in Fig. \ref{fig:cifar10treecompare} . 

Once the new classes have been assigned locations in the Tree-CNN, we begin the re-training of the network. The root node is re-trained using all 10 classes, divided into to subclasses. The branch node "animal" is retrained using training data from 6 classes,  3 old and 3 new added to it. Similarly, branch node "vehicles" is retrained with training data from 4 classes, 3 old, 1 new. 

\subsection{Sequentially Adding Multiple Classes (CIFAR-100)}

\begin{table}[h!]
\parbox[t]{0.48\linewidth}{
\centering
\caption{Root Node Tree-CNN (CIFAR-100)}
\label{table:Root_Node_Tree-CNN-C}
\scalebox{1}{
\begin{tabular}{|c|}
\hline
Input 32$\times$32$\times$3\\ \hline
CONV-1\\
64 5$\times$5 ReLU\\ \hline
{[}2 2{]} Max Pooling\\ \hline
CONV-2\\ 128 3$\times$3 ReLU\\ Dropout 0.5\\
128 3$\times$3 ReLU\\ \hline
{[}2 2{]} Max Pooling\\ \hline
CONV-3\\ 
256 3$\times$3 ReLU\\ Dropout 0.5\\
256 3$\times$3 ReLU\\ \hline
{[}2 2{]} Avg Pooling\\ \hline
FC\\ 
4096$\times$1024 ReLU\\ Dropout 0.5\\
1024$\times$1024 ReLU\\ Dropout 0.5\\
1024$\times$N\\ ($N = \#$ of Children)\\ \hline
\end{tabular}
}}
\hfill
\parbox[t]{0.48\linewidth}{
\centering
\caption{Branch Node Tree-CNN (CIFAR-100)}
\label{table:Branch_Node_Tree-CNN-C}
\scalebox{1}{
\begin{tabular}{|c|}
\hline
Input 32$\times$32$\times$3\\ \hline
CONV-1\\ 
32 5$\times$5 ReLU\\ \hline
{[}2 2{]} Max Pooling\\ \hline
Dropout 0.25\\ \hline
CONV-2\\
64 5$\times$5 ReLU \\ \hline
{[}2 2{]} Max Pooling \\ \hline
Dropout 0.25\\ \hline
CONV-3\\
64 3$\times$3 ReLU \\ 
Dropout 0.5\\ 
64 3$\times$3 ReLU \\ \hline
{[}2 2{]} Avg Pooling \\ \hline
FC\\
1024$\times$512 ReLU\\ Dropout 0.5\\
512$\times$128 ReLU\\ Dropout 0.5\\
128$\times$N\\ (N $= \#$ of Children)\\ \hline
\end{tabular}
}}
\end{table}

\subsubsection{Dataset}
The dataset, CIFAR-100 \cite{krizhevsky2009learning}, has $100$ classes, $500$ training and $100$ testing images per class. The $100$ classes are randomly divided into $10$ groups of 10 classes each and organized in a fixed order (\ref{appendix:incremental-cifar-100}). These groups of classes are introduced to the network incrementally.  

\subsubsection{The Network Initialization}

We initialize the \textit{Tree-CNN} as a root node with 10 leaf nodes. The root node, thus comprises of a CNN (Table \ref{table:Root_Node_Tree-CNN-C}), with 10 output nodes. Initially this CNN is trained to classify the 10 classes belonging to group 0 of the incremental CIFAR-100 dataset (\ref{appendix:incremental-cifar-100}). In subsequent learning stages, as new classes get grouped together under same output nodes, the network adds branch nodes. The DCNN model used in these branch nodes is given in Table \ref{table:Branch_Node_Tree-CNN-C}. The branch node has a higher chance of over-fitting than the root node as the dataset per node shrinks in size as we move deeper into the tree. Hence we introduce more dropout layers to the CNNs at these nodes to enhance regularization.

\subsubsection{Incremental Learning}
The remaining 9 groups, each containing 10 classes is incrementally introduced to the network in 9 learning stages. At each stage, 50 images belonging to each class are shown to the root node and a likelihood matrix $L$ is generated. The columns of the matrix are used to form an ordered set $S$, as described in section \ref{subsec:algorithm} . For this experiment, we applied the following constraints to the Algorithm \ref{algo:grow-tree}:
\begin{itemize}
\item Maximum depth of the tree is $2$. 
\item We set $\alpha = 0.1$ and $\beta = 0.1$. 
\item Maximum number of child nodes for a branch node is set at $5$, $10$, $20$ for the three test cases: \textit{Tree-CNN-5},\textit{Tree-CNN-10}, and \textit{Tree-CNN-20} respectively.
\end{itemize}

At every learning stage, once the new classes have been assigned the location in the Tree-CNN, we update the corresponding branch and root CNNs by retraining them on the combined dataset of old and new classes added to them. The branch nodes to which new children have not been added are left untouched. 

\subsection{Benchmarking}
\label{subsec:benchmarking}
There is an absence of standardized benchmark protocol for incremental learning, which led us to use a benchmarking protocol similar to one used in iCaRL \cite{rebuffi2017icarl}. The classes of the dataset are grouped and arranged in a fixed random order. At each learning stage, a selected set of classes would be introduced to the network. Once training is completed for a particular learning stage, the network would be evaluated on all the classes it has learned so far and the accuracy is reported.
\subsubsection{Baseline Network}
To compare against the proposed \textit{Tree-CNN}, we defined a baseline network (Network \textit{B}) with a complexity level similar to two stage \textit{Tree-CNN}. The network is has a VGG-net \cite{simonyan2014very} like structure with 11 layers. It has 4 convolutional blocks, each block having $2$ sets of $3\times3$ convolutional kernels (Table \ref{table:Network_B}). 

\subsubsection{Fine-tuning the baseline network using old $+$ new data}
The baseline network is trained in incremental stages using fine-tuning. The new classes are added as new output nodes of the final layer and $5$ different fine tuning strategies have been used. Each method retrains/fine-tunes certain layers of the network. While fine tuning, all of the available dataset is used, both old data and new data. It is assumed that the system has access to all the data that has been introduced so far. As listed below, we set $5$ different depths of back-propagation when retraining with the incremental data and the old data.
\begin{itemize}
\item B:I [FC]
\item B:II [FC $+$ CONV-1]
\item B:III [FC $+$ CONV-1 $+$ CONV-2]
\item B:IV [FC $+$ CONV-1 $+$ CONV-2 $+$ CONV-3]
\item B:V [FC $+$ CONV-1 $+$ CONV-2 $+$ CONV-3 $+$ CONV-4] (equivalent to training a new network with all the classes)
\end{itemize}

\subsubsection{Evaluation Metrics}
We compare \textit{Tree-CNN} against retraining Network \textit{B} on two metrics: Testing Accuracy, and Training Effort , which is defined as \newline
Training Effort $= \sum\limits_{nets}$ (total number of weights $\times$ total number of training samples)

Training Effort attempts to capture the number of weight updates that happen per training epoch. As batch size and number of training epochs is kept the same, the product of the number of weights and the number of training samples used can provide us with the measure of the computation cost of a learning stage. For \textit{Tree-CNN} the training effort of each of the nodes (or nets) is summed together. Whereas, for network $B$, it is just one node/neural network, and for each of the cases (\textit{B:I-B:V}), we simply sum the number of weights in the layers that are being retrained and multiply it with total number of training samples available at a learning stage to calculate the Training Effort. 

\subsection{The Training Framework}
We used MatConvNet \cite{Vedaldi_2015}, an open-source Deep Learning toolbox for MATLAB \cite{MATLAB_2017}, for training the networks. During training, data augmentation was done by flipping the  training images horizontally at random with a probability of $0.5$ \cite{goodfellow2013maxout}. All images were whitened and contrast normalized \cite{goodfellow2013maxout}. The activation used in all the networks is rectified linear activation ReLU, $\sigma(x) = max(x,0)$. The networks are trained using mini-batch stochastic gradient descent with fixed momentum of $0.9$. Dropout \cite{srivastava2014dropout} is used between the final fully connected layers, and between pooling layers to regularize the network. We also employed batch-normalization \cite{ioffe2015batch} at the output of every convolutional layer. Additionally, a weight decay $\lambda = 0.001$ was set to regularize each model. The weight decay helps against overfitting of our model. The final layer performs softmax operation on the output of the nodes to generate class probabilities. All CNNs are trained for $300$ epochs. The learning rate is kept at $0.1$ for first $200$ epochs, then reduced by a factor of $10$ every $50$ epochs.

\section{Results}\label{sec:results}

\subsection{Adding multiple new classes (CIFAR-10)} 

\begin{table}[h!]
\centering
\caption{Training Effort and Test Accuracy comparison of Tree-CNN against Network B for CIFAR-10}
\label{table:cifar10_te_test_accuracy}
\addtolength{\tabcolsep}{-3pt}
\begin{tabular}{|c|c|c|c|c|c|c|}
\hline  & B:I & B:II & B:III & B:IV & B:V  & Tree-CNN  \\ \hline
Testing Accuracy  & 78.37 & 85.02 & 88.15  & 90.00 & 90.51 & 86.24 \\ \hline
Normalized Training Effort & 0.40 & 0.85 & 0.96 & 0.99 & 1 & 0.60 \\ \hline
\end{tabular}
\end{table}

\begin{figure}[h!]

\begin{subfigure}[t]{0.48\linewidth}
\centering
\includegraphics[width=\linewidth]{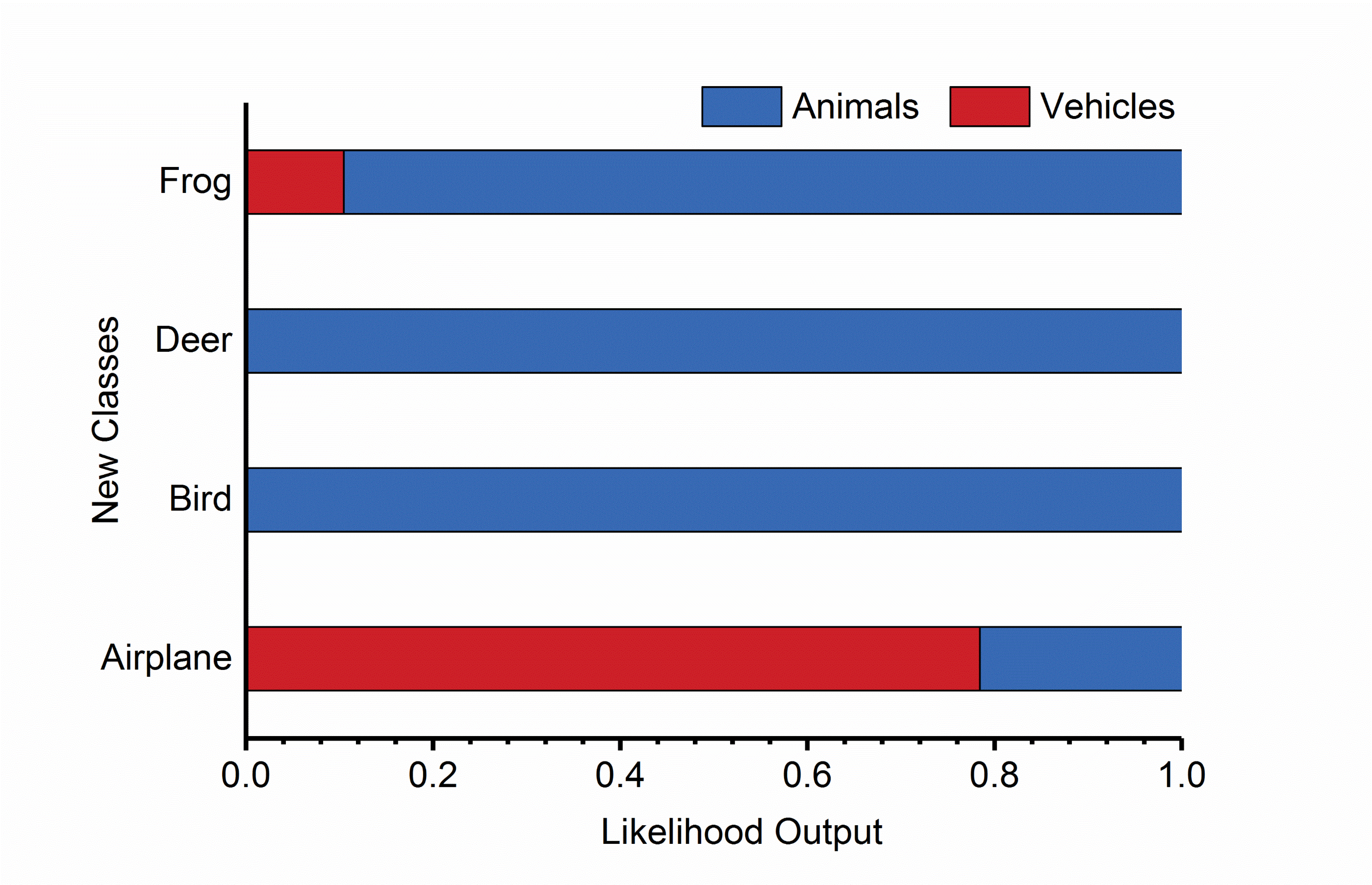}  
\caption{Likelihood values at the root node for the new classes}
\label{fig:cifar10_likelihood}
\end{subfigure}
\hfill
\begin{subfigure}[t]{0.48\linewidth}
\centering
\includegraphics[width=0.9\linewidth]{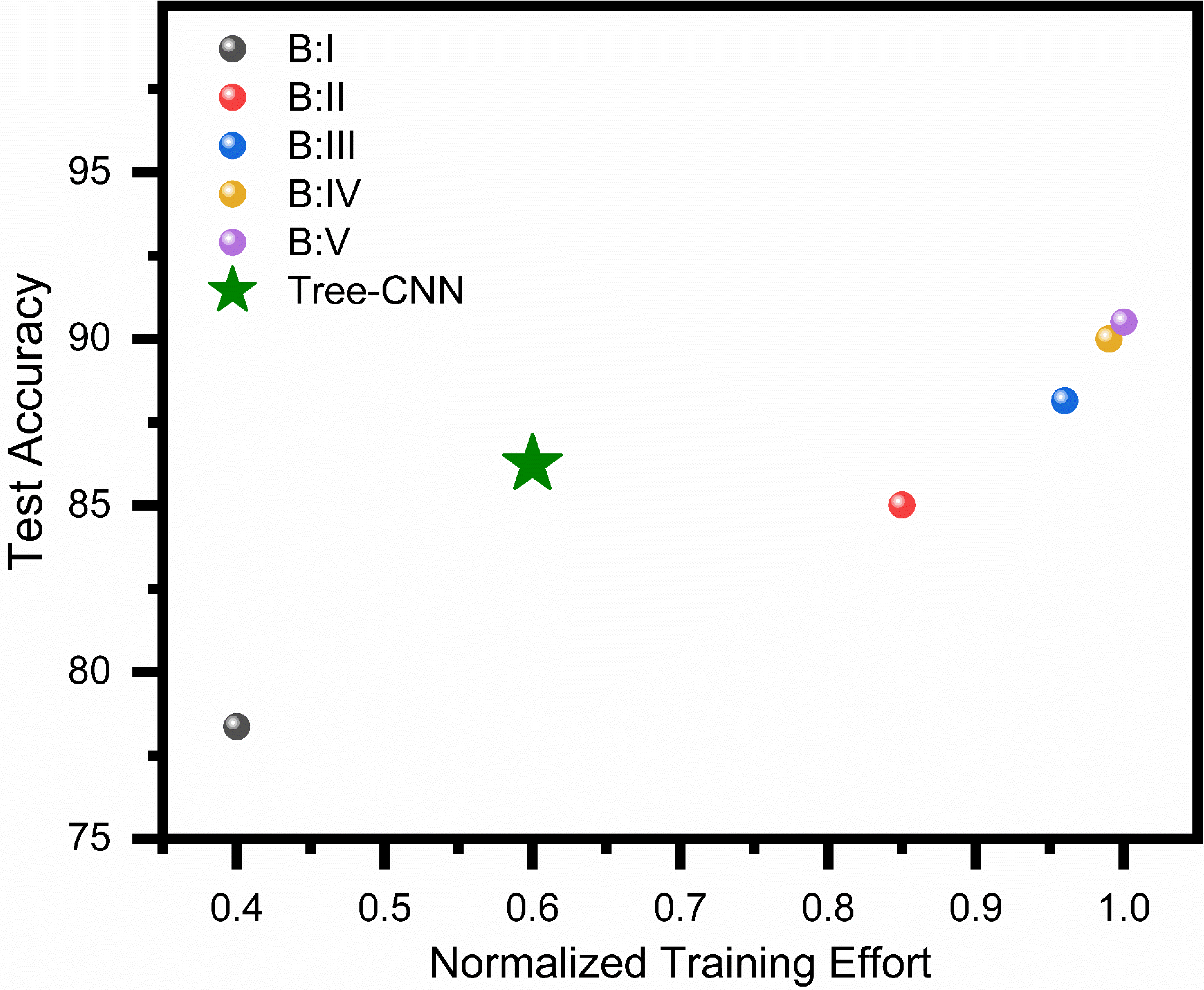}  
\caption{Testing Accuracy vs Normalized Training Effort (CIFAR-10)}
\label{fig:cifar10_tradeoff}
\end{subfigure} 
\caption{Incrementally learning CIFAR-10: 4 New classes are added to Network B and \textit{Tree-CNN}. Networks \textit{B:I} to \textit{B:V} represent 5 increasing depths of retraining for Network B.  (a) The softmax likelihood output at the root node for the two branches. (b) Testing Accuracy vs Normalized Training Effort for \textit{Tree-CNN} and networks \textit{B:I} to \textit{B:V} }
\end{figure}

We initialized a \textit{Tree-CNN} that can classify six classes (Fig. \ref{fig:cifar10treecompare}a). It had a root node and two branch nodes. The sample images from the 4 new classes generated the softmax likelihood output at root node as shown in Fig. \ref{fig:cifar10_likelihood}. Accordingly, the new classes are added to the two nodes, and the new \textit{Tree-CNN} is shown in Fig. \ref{fig:cifar10treecompare}b. In Table \ref{table:cifar10_te_test_accuracy}, we report the test accuracy and the training effort for the 5 cases of fine-tuning network \textit{B} against our \textit{Tree-CNN} for CIFAR-10. We observe that retraining only the FC layers of baseline network (\textit{B:I}) requires the least training effort, however, it gives us the lowest accuracy of $78.37\%$. And as more classes are introduced, this method causes significant loss in accuracy, as shown with CIFAR-100 (Fig. \ref{fig:cifar100_test_accuracy}). The \textit{Tree-CNN} has the second lowest normalized training effort, $\sim 40\%$ less than \textit{B:V}, and $\sim 30\%$ less than \textit{B:II}. At the same time, \textit{Tree-CNN} had comparable accuracy to \textit{B:II} and \textit{B:III}, while just being less than the ideal case \textit{B:V} by a margin of $3.76\%$. This accuracy vs training effort trade-off is presented in Fig. \ref{fig:cifar10_tradeoff}, where it is clearly visible that \textit{Tree-CNN} provided the most optimal solution for adding the 4 new classes.

\subsection{Sequentially adding new classes (CIFAR-100)} 

\begin{figure}[h]
\begin{subfigure}[t]{0.48\linewidth}
\includegraphics[width=\linewidth]{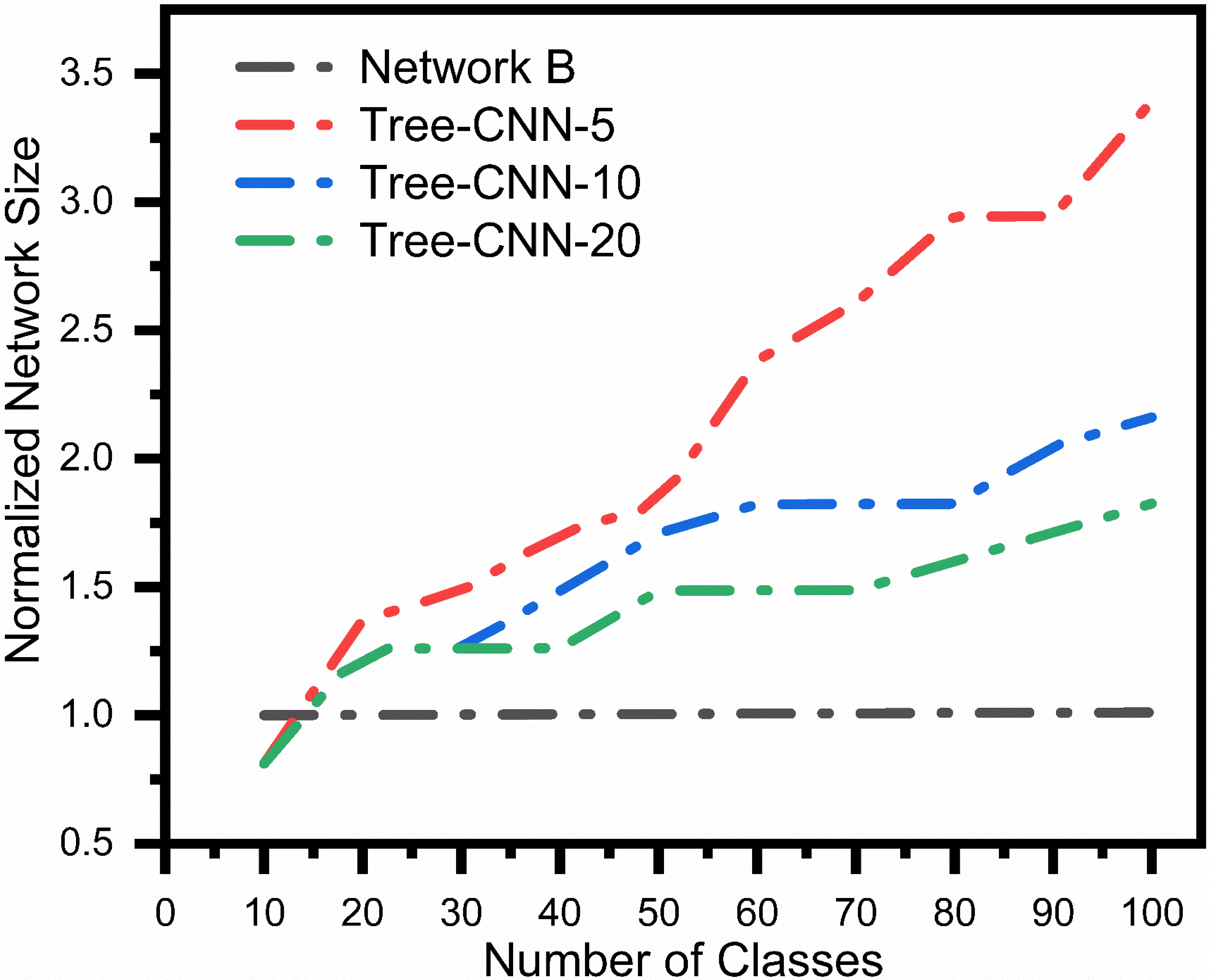}
\caption{Network size of the 3 \textit{Tree-CNNs} as new classes are added to the models, normalized with respect to Network \textit{B} (CIFAR-100)}
\label{fig:cifar100_network_size}
\end{subfigure}
\hfill
\begin{subfigure}[t]{0.48\linewidth}
\includegraphics[width=\linewidth]{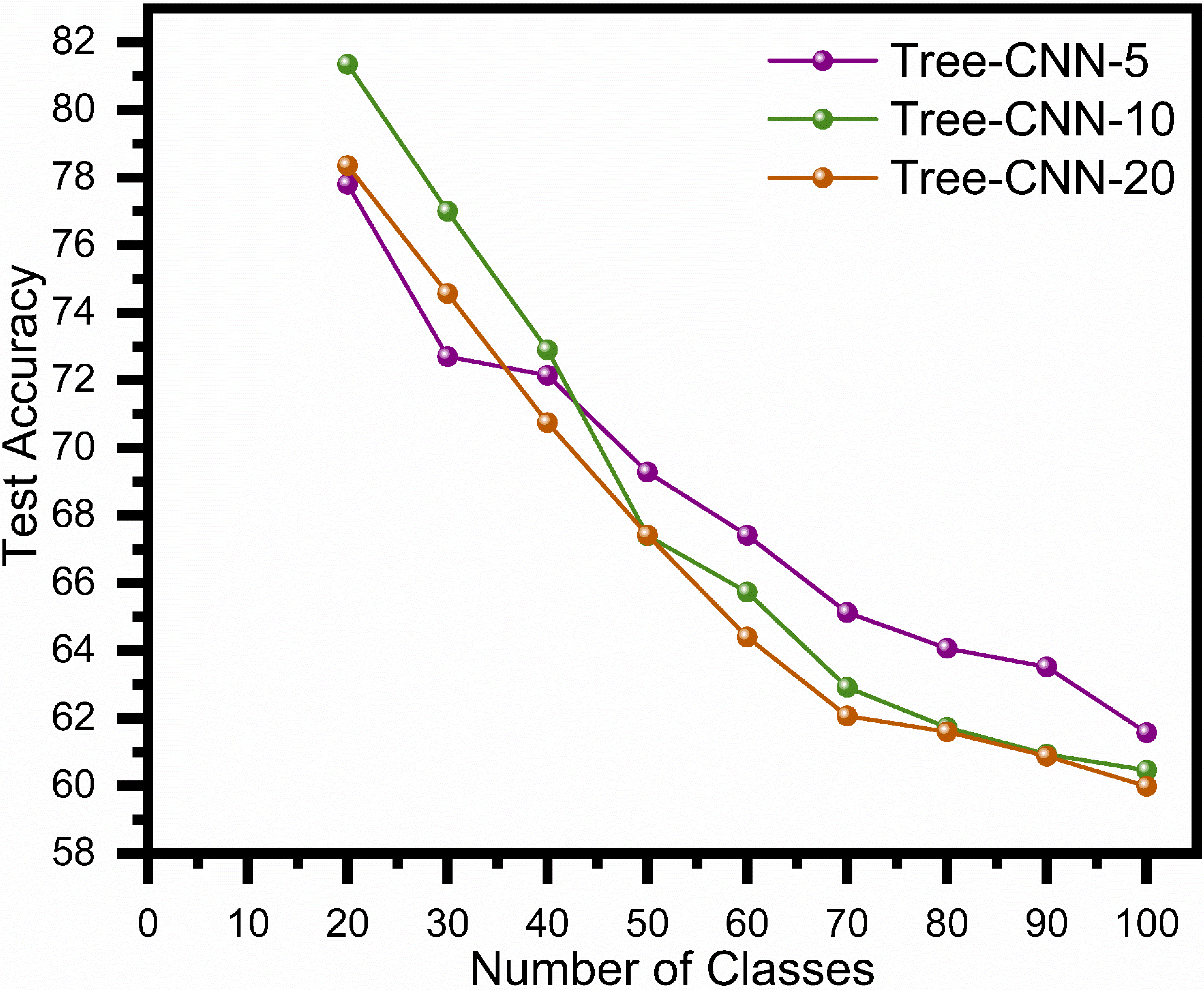}
\caption{Test Accuracy of the 3 \textit{Tree-CNNs} as as new classes are added to the models (CIFAR-100)}
\label{fig:cifar100_maxchild_acc}
\end{subfigure}
\caption{\textit{Tree-CNN}: Effect of varying the maximum number of children per branch node ($maxChildren$) as new classes are added to the models (CIFAR-100)}
\end{figure}

\begin{figure}[h]
\begin{subfigure}[t]{0.48\linewidth}
\centering
\includegraphics[width=0.9\linewidth]{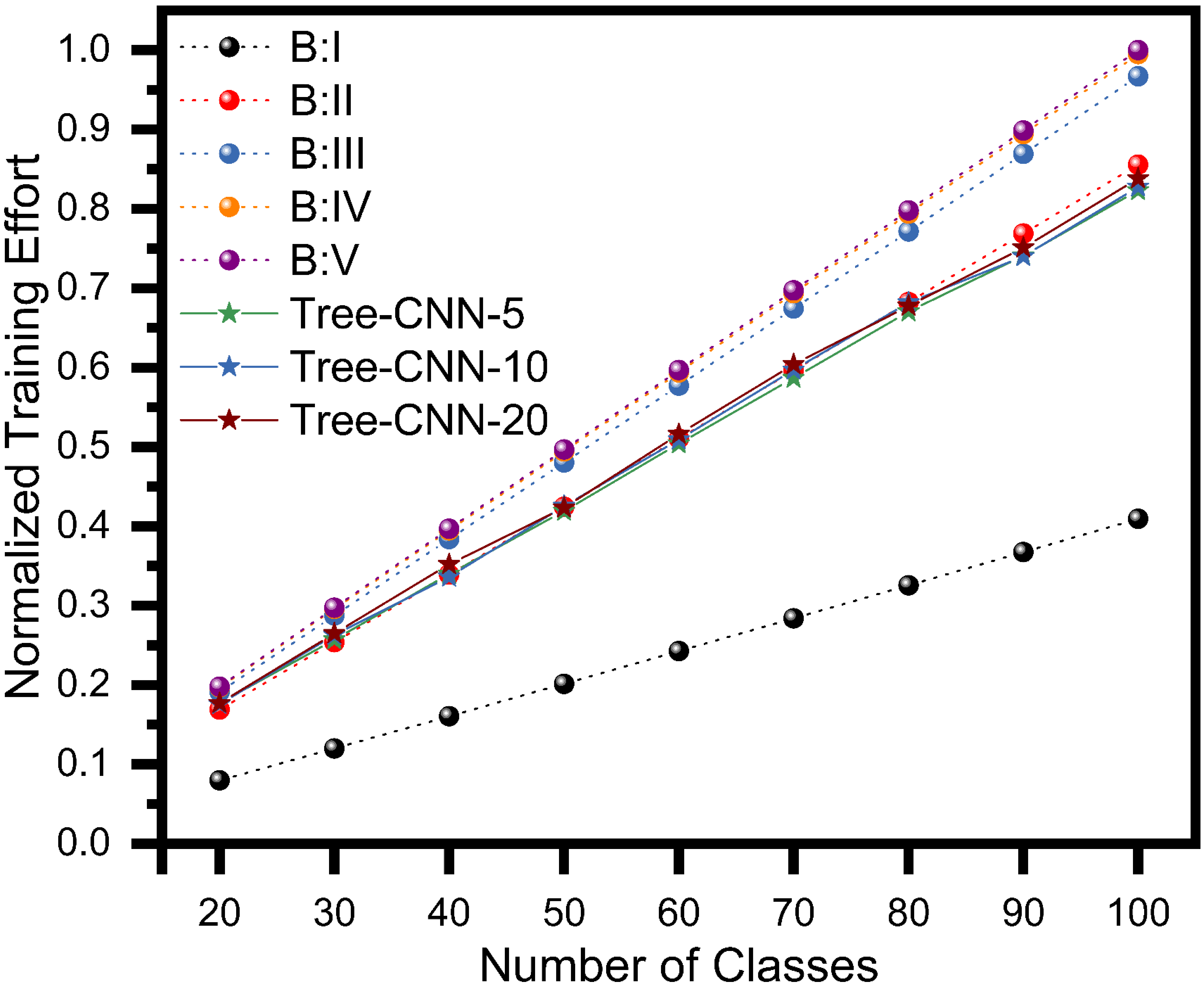}  
\caption{Training Effort (CIFAR-100) }
\label{fig:cifar100_te}
\end{subfigure}
\hfill
\begin{subfigure}[t]{0.48\linewidth}
\centering
\includegraphics[width=0.9\linewidth]{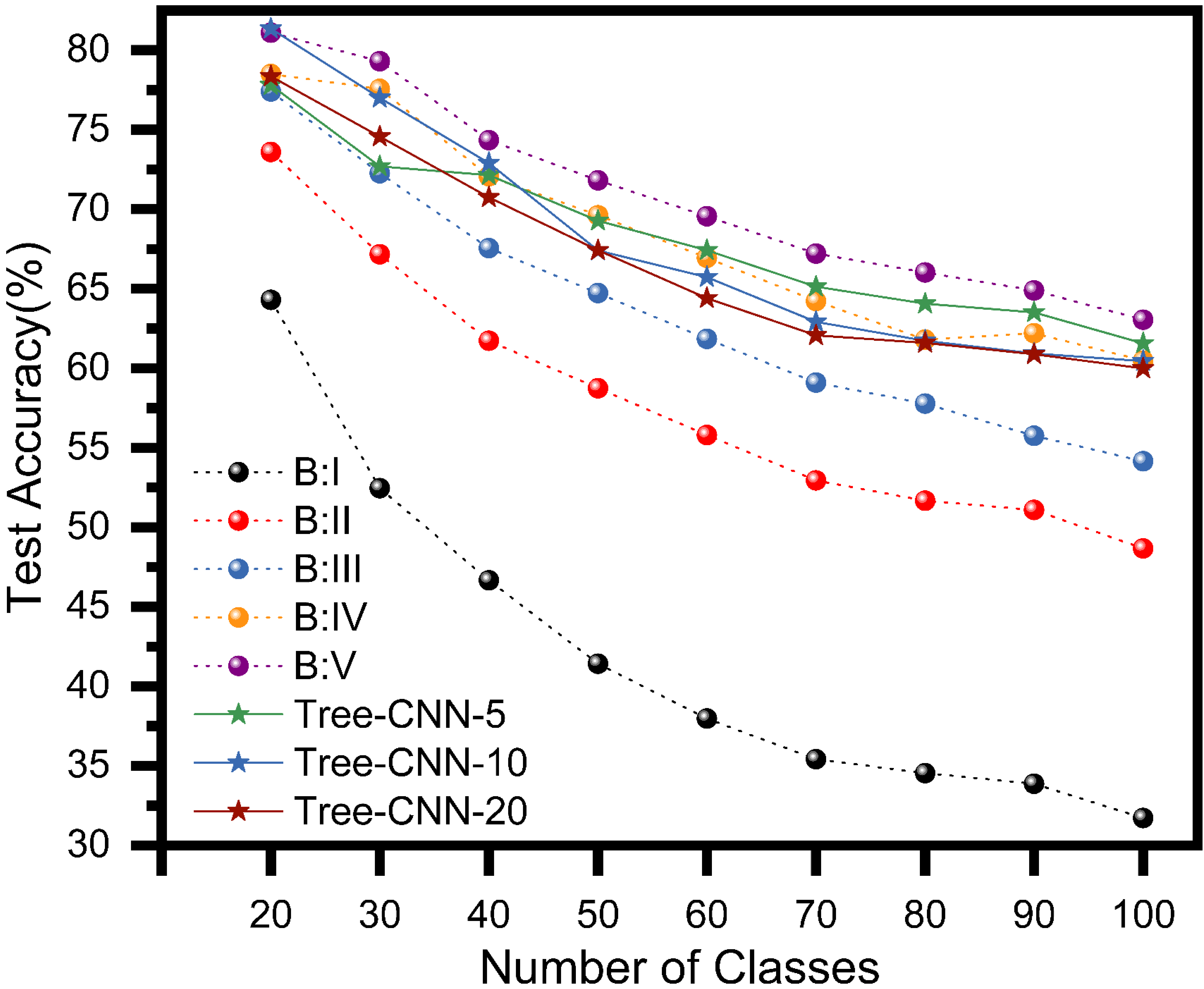}  
\caption{Testing Accuracy (CIFAR-100)}
\label{fig:cifar100_test_accuracy}
\end{subfigure} 
\caption{CIFAR-100: New classes are added to Network B and \textit{Tree-CNNs} in batches of 10. Networks B:I to B:V represent 5 increasing depths of retraining for Network B (a) Training effort for every learning stage (Table \ref{table:cifar100_training_effort}) (b) Testing Accuracy at the end of each learning stage (Table \ref{table:cifar100_testaccuracy})}
\label{fig:cifar100_te_acc}
\end{figure}

We initialized a root node that can classify 10 classes, i.e. has 10 leaf nodes. Then, we incrementally grew the \textit{Tree-CNN} for 3 different values of maximum children per branch node ($maxChildren$), namely 5, 10, and 20. We label these 3 models as \textit{Tree-CNN-5}, \textit{Tree-CNN-10} and \textit{Tree-CNN-20} respectively. At the end of 9 incremental learning stages, the root node of \textit{Tree-CNN-5} had 23 branch nodes and 3 leaf nodes. Whereas, the root node of \textit{Tree-CNN-10} has 12 branch nodes and 5 leaf nodes. As expected, the root node of \textit{Tree-CNN-20} had least number of child nodes, 9 branch nodes and 3 leaf nodes. The final hierarchical structure of the \textit{Tree-CNNs} can be found in \ref{appendix:tree_cnn_final_5_10_20}, Fig. \ref{fig:tree_cnn_5_final}-\ref{fig:tree_cnn_20_final}. 

We observe that as new classes are added, the \textit{Tree-CNNs} grow in size by adding more branches (Fig. \ref{fig:cifar100_network_size}). The size of Network \textit{B} remains relatively unchanged, as only additional output nodes are added, which translates to a small fraction of new weights in the final layer. \textit{Tree-CNN-5} almost grows $3.4\times$ the size of Network \textit{B}, while \textit{Tree-CNN-10} and \textit{Tree-CNN-20} reach $2.2\times$ and $1.8\times$ the baseline size, respectively. The training effort for the 3 \textit{Tree-CNNs} was almost identical, within 1e-2 margin of each other (Fig. \ref{fig:cifar100_te}), over the 9 incremental learning stages.  As $maxChildren$ is reduced, the test accuracy improves, as observed in Fig. \ref{fig:cifar100_maxchild_acc}. If $maxChildren$ is set to 1, we obtain a situation similar to test case \textit{B:V}, where every new class is just a new output node.  

We compare the training effort needed for the \textit{Tree-CNNs} against the 5 different fine-tuning cases of Network \textit{B} over the 9 incremental learning stages in Fig. \ref{fig:cifar100_te}. We normalized the training effort by dividing all the values with the highest training effort. i.e. \textit{B:V}. For all the models, the training effort required at a particular learning stage was greater than the effort required by the previous stage. This is because we had to show images belonging to old classes to avoid ``catastrophic forgetting''. The \textit{Tree-CNNs} exhibit a lower training effort than 4 fine-tuning test cases, \textit{B:II - B:V}. the test case \textit{B:I} has a significantly lower training effort than all the other cases, as it only retrains the final fully connected layer. However, it suffers the worst accuracy degradation over the 9 learning stages (Fig. \ref{fig:cifar100_test_accuracy}). This shows that only retraining the final linear classifier, i.e. the fully connected layer, is not sufficient. We need to train the feature extractors, i.e. convolutional blocks, as well on the new data. 

\begin{figure}[h!]
\begin{subfigure}[t]{0.48\linewidth}
\centering
\includegraphics[width=\linewidth]{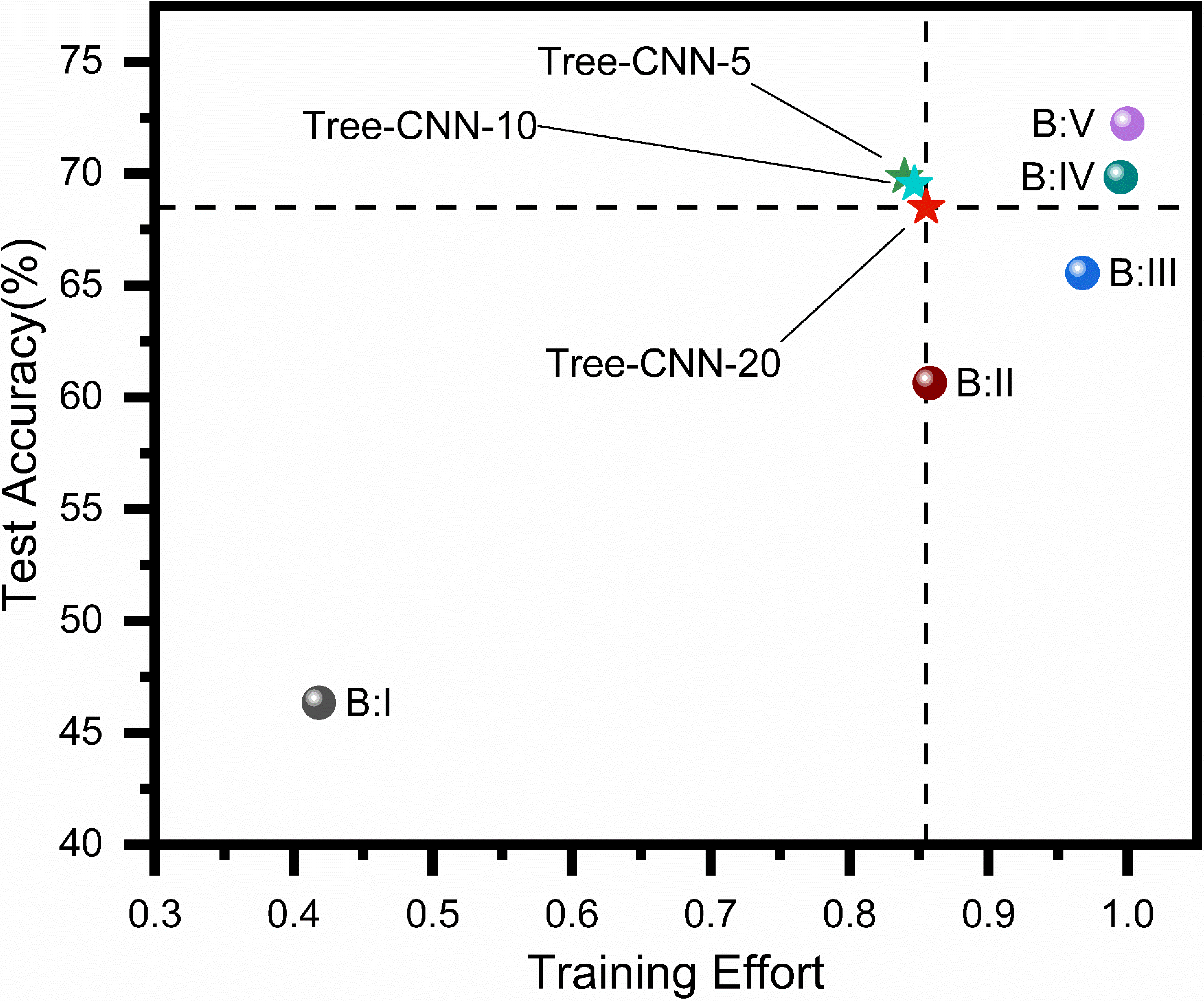}  
\caption{Average test accuracy vs average training effort over all the learning stages of \textit{Tree-CNNs} and Network \textit{B:I - B:V} (CIFAR-100)}  
\label{fig:cifar100_tradeoff}
\end{subfigure}
\hfill
\begin{subfigure}[t]{0.48\linewidth}
\centering
\includegraphics[width=\linewidth]{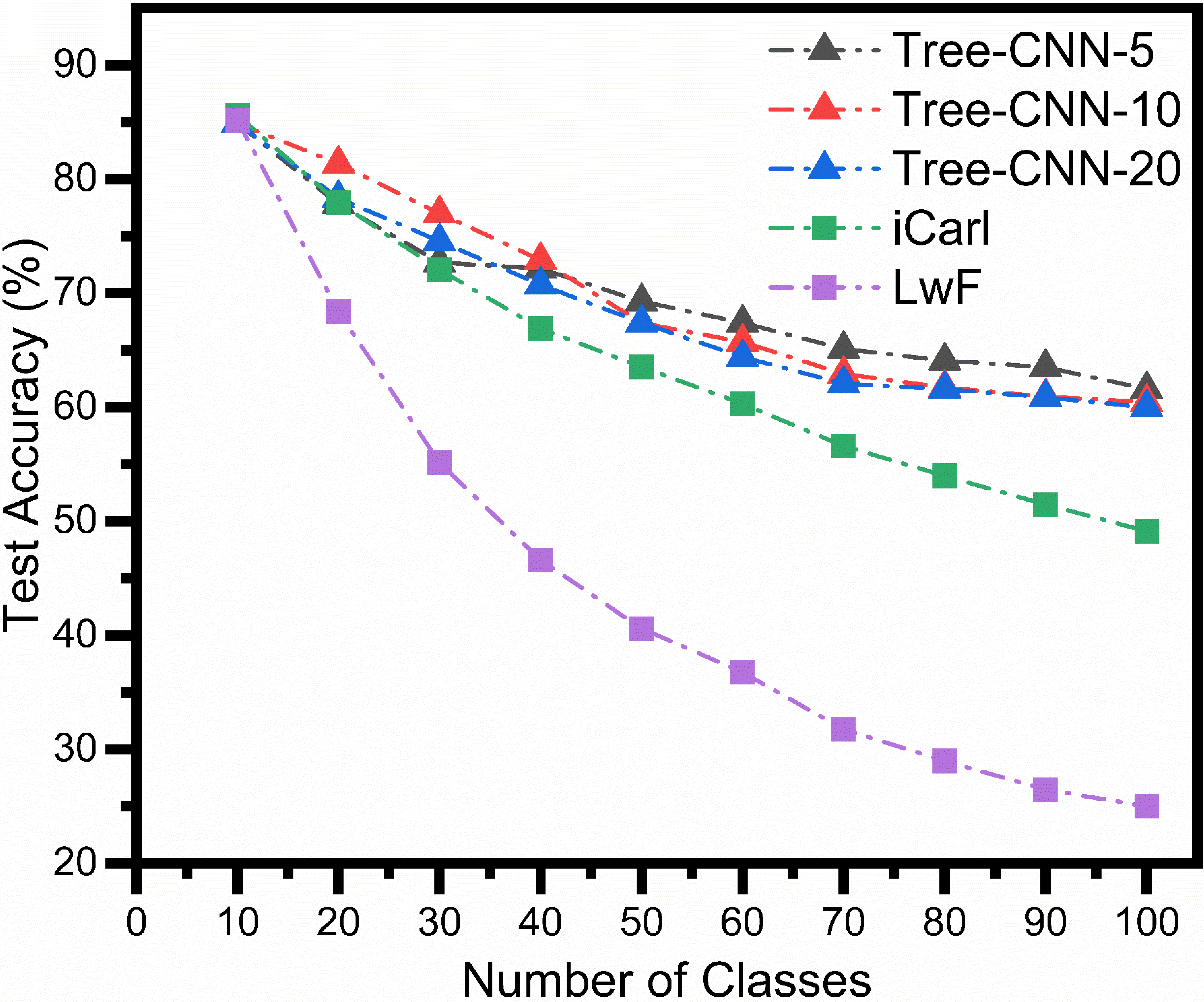}  
\caption{Accuracy over incremental learning stages of \textit{Tree-CNNs}, iCaRL\cite{rebuffi2017icarl} and Learning without Forgetting(LwF) \cite{li2017learning} where new classes are added in batches of 10 (CIFAR-100)}  
\label{fig:cifar100_compare}
\end{subfigure}
\caption{The performance of \textit{Tree-CNN} compared with a) Fine-tuning Network \textit{B} b) Other incremental learning methods \cite{rebuffi2017icarl,li2017learning}}
\label{fig:cifar100_tradeoff_compare}
\end{figure}

While \textit{B:I} is the worst performer in terms of accuracy, Fig. \ref{fig:cifar100_test_accuracy} shows that all the networks suffer from some accuracy degradation with increasing number of classes. \textit{B:V} provides the baseline accuracy at each stage, as it represents a network fully trained on all the available data at that stage. The three \textit{Tree-CNNs} perform almost at par with \textit{B:IV}, and outperform all other variants of network \textit{B}. From Fig. \ref{fig:cifar100_te_acc}, we can conclude that \textit{Tree-CNNs} offer the most optimal trade-off between training effort and testing accuracy. This is further illustrated in Fig. \ref{fig:cifar100_tradeoff}, where we plot the average test accuracy and average training effort over all the learning stages. 

\begin{table}[h!]
\centering
\caption{Test Accuracy over all 100 classes of CIFAR-100}
\label{table:cifar100_final_accuracy}
\begin{tabular}{|c|C{3 cm}|C{3 cm}|}
\hline
Model                     & Final Test Accuracy (\%) & Average Test Accuracy(\%) \\ \hline
B:V                       & 63.05                    & 72.23 \\ \hline
Tree-CNN-5                & 61.57                    & 69.85 \\ \hline
Tree-CNN-10               & 60.46                    & 69.53 \\ \hline
Tree-CNN-20               & 59.99                    & 68.49 \\ \hline
iCarl (Rebuffi, et al. 2017) \cite{rebuffi2017icarl}         & 49.11 & 64.10 \\ \hline
LwF (Li, et al. 2017) \cite{rebuffi2017icarl,li2017learning} & 25.00 & 44.49 \\ \hline
HD-CNN (Yan, et al. 2015) \cite{Yan_2015} & 67.38  & N/A                   \\ \hline
Hertel, et al. 2015  \cite{Hertel_2015}   & 67.68  & N/A                   \\ \hline
\end{tabular}
\end{table}

We compare our model against two works on incremental learning, `iCaRL'\cite{rebuffi2017icarl} and `Learning without Forgetting' \cite{li2017learning} as shown in Fig. \ref{fig:cifar100_compare}. We use the accuracy reported in \cite{rebuffi2017icarl} for CIFAR-100, and compare it against our method. For `LwF', a ResNet-32 \cite{he2016deep} is retrained exclusively on new data at every stage. Hence it suffers the most accuracy degradation. In `iCarl' \cite{rebuffi2017icarl}, a ResNet-32 is retrained  with new data and only 2000 samples of old data (called `exemplars') at every stage. It is able to recover a good amount performance, compared to `LwF' but still falls short of state-of-the-art by $\approx$18\%. \textit{Tree-CNNs} yield 10\% higher accuracy than `iCaRL' and over 50\% higher accuracy than `Learning without Forgetting' (LwF). This shows that our learning method using the hierarchical structure is more resistant to catastrophic forgetting as new classes are added. \textit{Tree-CNNs} are able to achieve near state-of-the-art accuracy for CIFAR-100 as illustrated in Table \ref{table:cifar100_final_accuracy}. While the second column reports the final accuracy, the third column reports the average accuracy of the incremental learning methods where new classes are added in batches of 10. 

\begin{figure}[h!]
\centering
\includegraphics[width=0.6\linewidth]{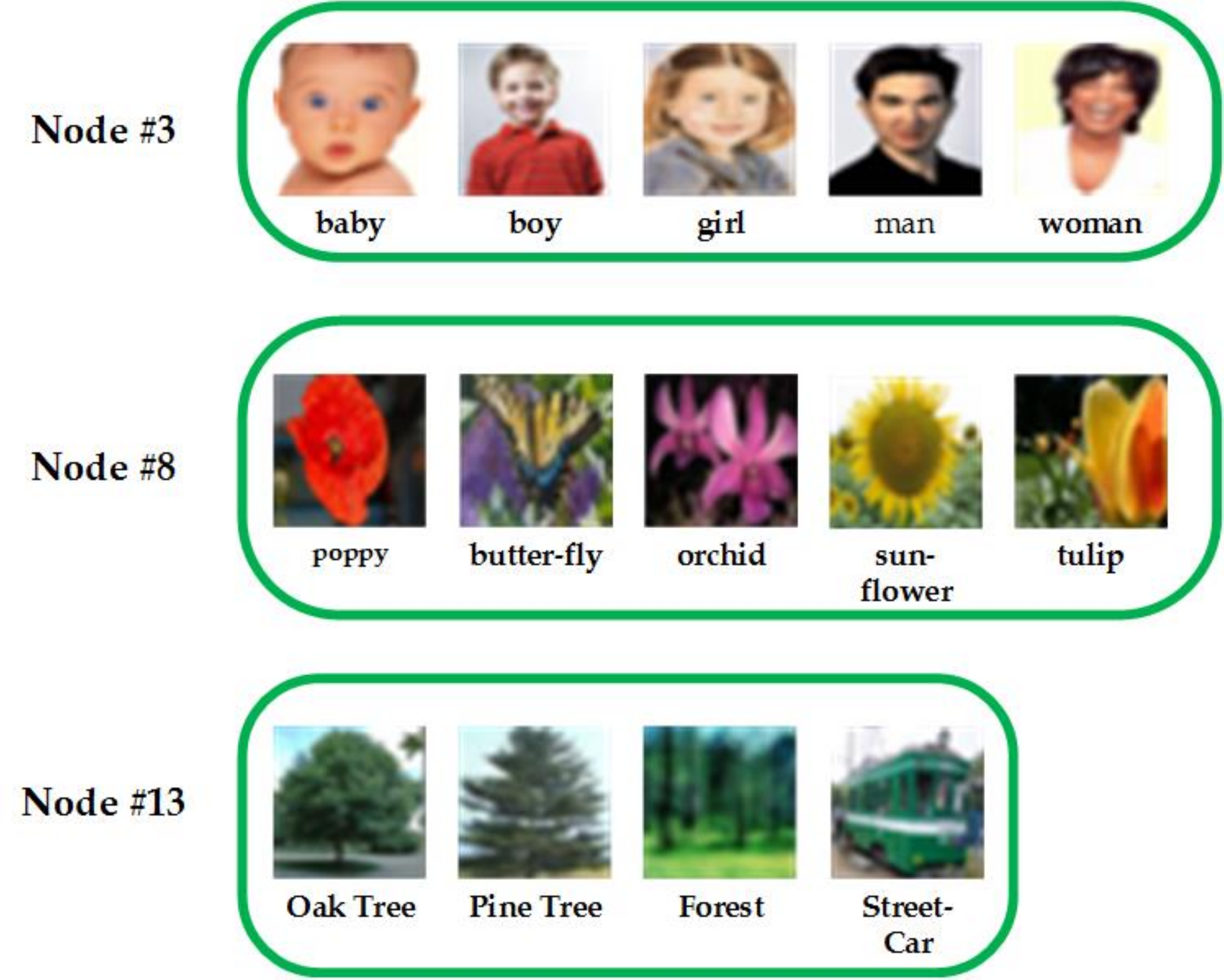}
  \caption{Examples of groups of classes formed when new classes were added incrementally to \textit{Tree-CNN-10} in batches of 10 for CIFAR-100}
\label{fig:cifar100examples}
\end{figure}

An interesting thing to note was similar looking classes, that were also semantically similar, were grouped under the same branches. At the end of the nine incremental learning stages, certain similar objects grouped together is shown in Fig. \ref{fig:cifar100examples} for \textit{Tree-CNN-10}. While there were some groups that had object sharing semantic similarity as well, there were odd groups as well, such as Node 13 as shown in Fig. \ref{fig:cifar100examples}. This opens up the possibility of using such a hierarchical structure for finding hidden similarity in the incoming data. 

\section{Discussion}
\label{sec:discussion}
The motivation of this work stems from the idea that subsequent addition of new image classes to a network should be easier than retraining the whole network again with all the classes. We observed that each incremental learning stage required more effort than the previous, because images belonging to old
classes needed to be shown to the CNNs. This is due to the inherent problem of “catastrophic forgetting” in deep neural networks. Our proposed method offers the best trade-off between accuracy and training effort when compared against fine-tuning layers of a deep network. It also achieves better accuracy, much closer to state-of-the-art on CIFAR-100, as compared to other works, `iCarl' \cite{rebuffi2017icarl} and `LwF' \cite{li2017learning}. The hierarchical node-based learning model of the \textit{Tree-CNN} attempts to confine the change in the model to a few nodes only. And, in this way, it limits the computation costs of retraining, while using all the previous data. Thus, it can learn with lower training effort than fine-tuning a deep network, while preserving much of the accuracy. However, the \textit{Tree-CNN} continues to grow in size over time, and the implications of that on memory requirements needs to be investigated. During inference, a single node is evaluated at a time, thus the memory requirement per node inference is much lower than the size of the entire model. \textit{Tree-CNN} grows in a manner such that images that share common features are grouped together. The correlation of the semantic similarity of the class labels and the feature-similarity of the class images under a branch is another interesting area to explore. The \textit{Tree-CNN} generates hierarchical grouping of initially unrelated classes, thereby generating a label relation graph out of these classes[33]. The final leaf nodes, and the distance between them can also be used as a measure of how similar any two images are. Such a method of training and classification can be used to hierarchically classify large datasets. Our proposed method, Tree-CNN, thus offers a better learning model that is based on hierarchical classifiers and transfer learning and can organically adapt to new information over time.

\section*{Acknowledgment}

This work was supported in part by the Center for Brain Inspired Computing (C-BRIC), one of the six centers in JUMP, a Semiconductor Research Corporation (SRC) program sponsored by DARPA, the National Science Foundation, Intel Corporation, the DoD Vannevar Bush Fellowship, and by the U.S. Army Research Laboratory and the U.K. Ministry of Defense under Agreement Number W911NF-16-3-0001.

\section*{Declaration of Interests}
Declarations of interest: none

\pagebreak
\appendix
\section{Incremental CIFAR-100 Dataset}
\label{appendix:incremental-cifar-100}
The $100$ classes of CIFAR-100 were randomly arranged and divided in $10$ batches, each containing $10$ classes. We randomly shuffled numbers 1 to 100 in 10 groups and then used that to group classes. We list the batches in the order they were added to the Tree-CNN for the incremental learning task below.
\begin{itemize}
    \item[0] chair, bridge, girl, kangaroo, lawn mower, possum, otter, poppy, sweet pepper, bicycle
    \item[1] lion, man, palm tree,	tank, willow tree, bowl, mountain, hamster, chimpanzee, cloud
    \item[2] plain, leopard, castle, bee, raccoon,	bus,	rabbit,	train,	worm,	ray
    \item[3] table, aquarium fish,	couch, caterpillar, whale, sunflower, trout, butterfly, shrew, house
    \item[4] bottle, orange, dinosaur,	beaver,	bed, snail, flatfish, shark, tractor, apple
    \item[5] woman, fox, lobster, skunk, can, turtle, cockroach, dolphin, bear, pickup truck
    \item[6] lizard, road, porcupine, mouse, seal, sea, tiger, telephone, rocket, tulip
    \item[7] baby, motorcycle, elephant, clock, maple tree, mushroom, pear, orchid, spider, oak tree
    \item[8] wardrobe, squirrel, crocodile, wolf, plate, skyscraper, keyboard, beetle, streetcar, crab
    \item[9] snake, lamp, camel, pine tree, cattle, boy, rose, forest,	television,	cup
\end{itemize}

\pagebreak 

\section{Final Tree-CNN for max children 5, 10, 20 (CIFAR-100)}
\label{appendix:tree_cnn_final_5_10_20}
\begin{figure}[h]
\centering
\includegraphics[width=0.9\linewidth]{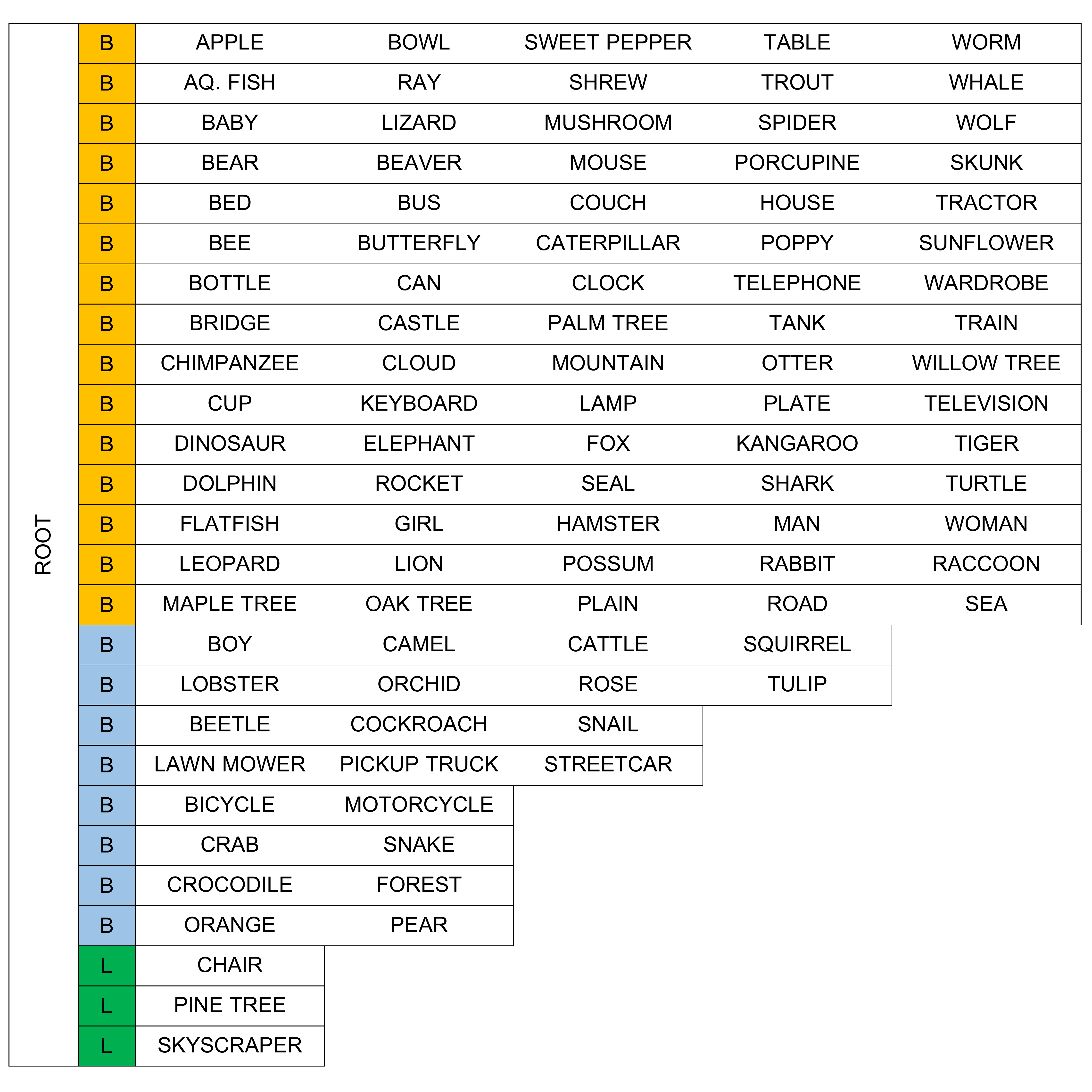}  
\caption{Tree-CNN-5: After 9 incremental learning stages}
\label{fig:tree_cnn_5_final}
\end{figure}

\begin{figure}[h]
\centering
\includegraphics[width=\linewidth]{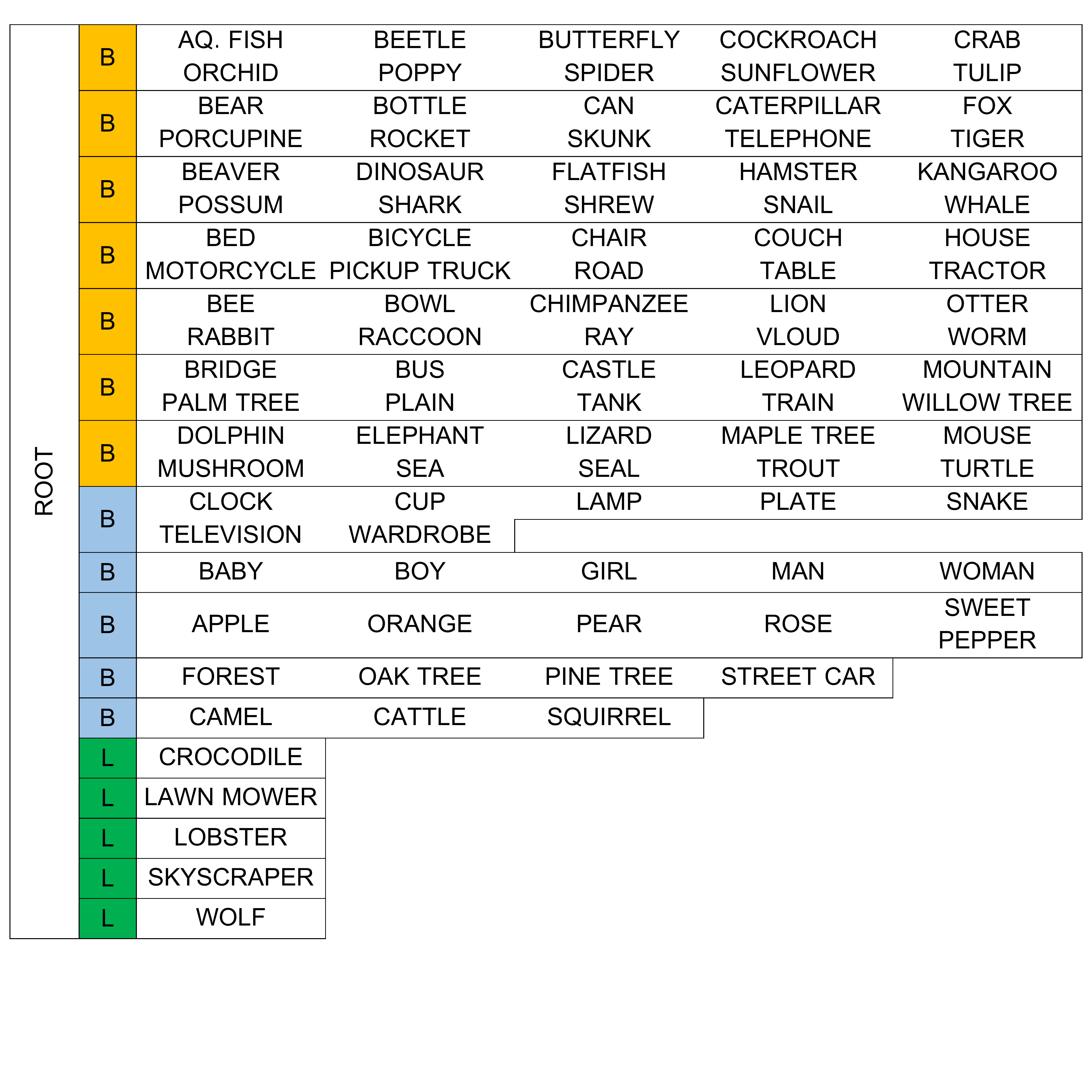}  
\caption{Tree-CNN-10: After 9 incremental learning stages}
\label{fig:tree_cnn_10_final}
\end{figure}

\begin{figure}[h]
\centering
\includegraphics[width=\linewidth]{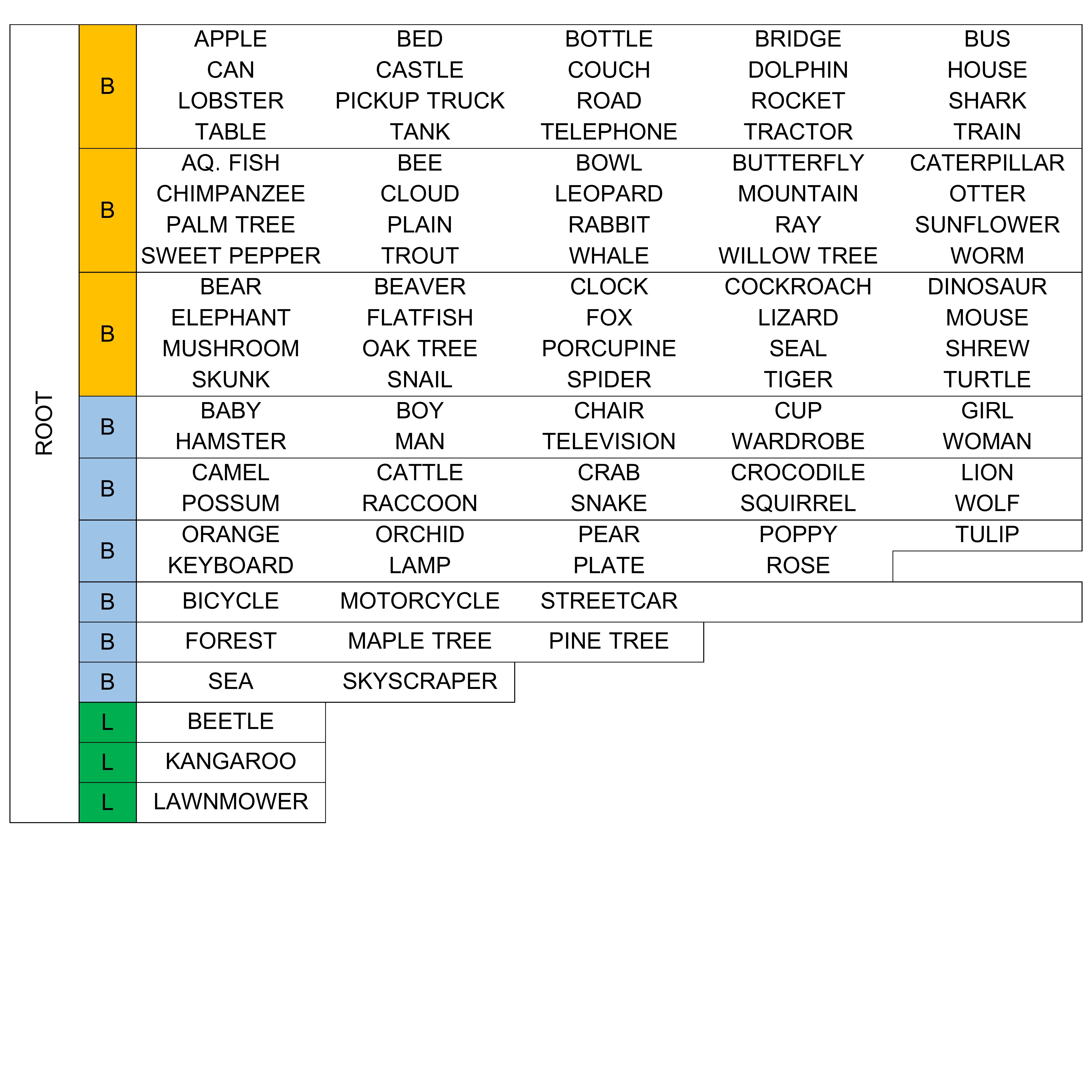}  
\caption{Tree-CNN-20: After 9 incremental learning stages}
\label{fig:tree_cnn_20_final}
\end{figure}

We trained the \textit{Tree-CNN} with the incremental CIFAR-100 dataset, and we set the maximum number of children a branch node can have as 5, 10, and 20. The corresponding 3 \textit{Tree-CNN}s were labeled - \textit{Tree-CNN-5}, \textit{Tree-CNN-10}, \textit{Tree-CNN-20}. The 2-level hierarchical structure of these \textit{Tree-CNNs} after 9 incremental learning stages is shown in Fig. \ref{fig:tree_cnn_5_final} - \ref{fig:tree_cnn_20_final}. The nodes marked `yellow' indicate completely filled branch nodes (B), the ones marked `blue' indicate branch nodes (B) that are partially filled, while those marked `green' refer to leaf nodes (L). 

\newpage

\newpage

\newpage
\section{Full Simulation Results}
\label{appendix:additional_results}

\begin{table}[h!]
\caption{Normalized Training Effort as classes are added incrementally in batches of 10 (CIFAR-100)}
\label{table:cifar100_training_effort}
\begin{tabular}{|C{2 cm}|c|c|c|c|c|C{1.5 cm}|C{1.5 cm}|C{1.5 cm}|}
\hline
Number of Classes & B:I  & B:II & B:III & B:IV & B:V  & Tree-CNN-5 & Tree-CNN-10 & Tree-CNN-20 \\ \hline
20                & 0.08 & 0.17 & 0.19  & 0.20 & 0.20 & 0.18       & 0.18        & 0.18        \\ \hline
30                & 0.12 & 0.25 & 0.29  & 0.30 & 0.30 & 0.26       & 0.26        & 0.27        \\ \hline
40                & 0.16 & 0.34 & 0.38  & 0.39 & 0.40 & 0.34       & 0.34        & 0.35        \\ \hline
50                & 0.20 & 0.42 & 0.48  & 0.49 & 0.50 & 0.42       & 0.43        & 0.42        \\ \hline
60                & 0.24 & 0.51 & 0.58  & 0.59 & 0.60 & 0.50       & 0.51        & 0.52        \\ \hline
70                & 0.28 & 0.60 & 0.67  & 0.69 & 0.70 & 0.59       & 0.60        & 0.60        \\ \hline
80                & 0.33 & 0.68 & 0.77  & 0.79 & 0.80 & 0.67       & 0.68        & 0.68        \\ \hline
90                & 0.37 & 0.77 & 0.87  & 0.89 & 0.90 & 0.74       & 0.74        & 0.75        \\ \hline
100               & 0.41 & 0.86 & 0.97  & 1.00 & 1.00 & 0.82       & 0.83        & 0.84        \\ \hline
\end{tabular}
\end{table}

\begin{table}[h!]
\caption{Test Accuracy as classes are added incrementally in batches of 10 (CIFAR-100)}
\label{table:cifar100_testaccuracy}
\begin{tabular}{|C{2 cm}|c|c|c|c|c|C{1.5 cm}|C{1.5 cm}|C{1.5 cm}|}
\hline
Number of Classes & B:I   & B:II  & B:III & B:IV  & B:V   & Tree-CNN-5 & Tree-CNN-10 & Tree-CNN-20 \\ \hline
20                & 64.30 & 73.60 & 77.40 & 78.50 & 81.10 & 77.80      & 81.35       & 78.35       \\ \hline
30                & 52.47 & 67.17 & 72.27 & 77.57 & 79.30 & 72.70      & 77.00       & 74.57       \\ \hline
40                & 46.68 & 61.72 & 67.55 & 72.08 & 74.35 & 72.15      & 72.90       & 70.75       \\ \hline
50                & 41.42 & 58.74 & 64.74 & 69.62 & 71.82 & 69.28      & 67.40       & 67.42       \\ \hline
60                & 37.98 & 55.80 & 61.85 & 66.95 & 69.57 & 67.42      & 65.73       & 64.40       \\ \hline
70                & 35.43 & 52.96 & 59.10 & 64.23 & 67.21 & 65.13      & 62.91       & 62.07       \\ \hline
80                & 34.55 & 51.68 & 57.77 & 61.81 & 66.03 & 64.07      & 61.73       & 61.60       \\ \hline
90                & 33.88 & 51.09 & 55.77 & 62.21 & 64.90 & 63.52      & 60.93       & 60.88       \\ \hline
100               & 31.73 & 48.68 & 54.16 & 60.48 & 63.05 & 61.57      & 60.46       & 59.99       \\ \hline
\end{tabular}
\end{table}

\newpage
\bibliographystyle{elsarticle-num-names} 
\bibliography{ref_list.bib}

\begin{thebibliography}{32}
\providecommand{\natexlab}[1]{#1}
\providecommand{\url}[1]{\texttt{#1}}
\providecommand{\urlprefix}{URL }
\expandafter\ifx\csname urlstyle\endcsname\relax
  \providecommand{\doi}[1]{doi:\discretionary{}{}{}#1}\else
  \providecommand{\doi}[1]{doi:\discretionary{}{}{}\begingroup
  \urlstyle{rm}\url{#1}\endgroup}\fi
\providecommand{\bibinfo}[2]{#2}

\bibitem[{Rawat and Wang(2017)}]{Rawat2017}
\bibinfo{author}{W.~Rawat}, \bibinfo{author}{Z.~Wang}, \bibinfo{title}{{Deep
  Convolutional Neural Networks for Image Classification: A Comprehensive
  Review}}, \bibinfo{journal}{Neural Computation}
  \bibinfo{volume}{29}~(\bibinfo{number}{9}) (\bibinfo{year}{2017})
  \bibinfo{pages}{2352--2449}.

\bibitem[{Krizhevsky et~al.(2012)Krizhevsky, Sutskever, and
  Hinton}]{krizhevsky2012imagenet}
\bibinfo{author}{A.~Krizhevsky}, \bibinfo{author}{I.~Sutskever},
  \bibinfo{author}{G.~E. Hinton}, \bibinfo{title}{Imagenet classification with
  deep convolutional neural networks}, in: \bibinfo{booktitle}{Advances in
  neural information processing systems}, \bibinfo{pages}{1097--1105},
  \bibinfo{year}{2012}.

\bibitem[{LeCun et~al.(1998)LeCun, Bottou, Bengio, and
  Haffner}]{lecun1998gradient}
\bibinfo{author}{Y.~LeCun}, \bibinfo{author}{L.~Bottou},
  \bibinfo{author}{Y.~Bengio}, \bibinfo{author}{P.~Haffner},
  \bibinfo{title}{Gradient-based learning applied to document recognition},
  \bibinfo{journal}{Proceedings of the IEEE}
  \bibinfo{volume}{86}~(\bibinfo{number}{11}) (\bibinfo{year}{1998})
  \bibinfo{pages}{2278--2324}.

\bibitem[{Wan et~al.(2013)Wan, Zeiler, Zhang, Cun, and
  Fergus}]{wan2013regularization}
\bibinfo{author}{L.~Wan}, \bibinfo{author}{M.~Zeiler},
  \bibinfo{author}{S.~Zhang}, \bibinfo{author}{Y.~L. Cun},
  \bibinfo{author}{R.~Fergus}, \bibinfo{title}{Regularization of neural
  networks using dropconnect}, in: \bibinfo{booktitle}{Proceedings of the 30th
  international conference on machine learning (ICML-13)},
  \bibinfo{pages}{1058--1066}, \bibinfo{year}{2013}.

\bibitem[{Krizhevsky and Hinton(2009)}]{krizhevsky2009learning}
\bibinfo{author}{A.~Krizhevsky}, \bibinfo{author}{G.~Hinton},
  \bibinfo{title}{Learning multiple layers of features from tiny images} .

\bibitem[{Russakovsky et~al.(2015)Russakovsky, Deng, Su, Krause, Satheesh, Ma,
  Huang, Karpathy, Khosla, Bernstein et~al.}]{russakovsky2015imagenet}
\bibinfo{author}{O.~Russakovsky}, \bibinfo{author}{J.~Deng},
  \bibinfo{author}{H.~Su}, \bibinfo{author}{J.~Krause},
  \bibinfo{author}{S.~Satheesh}, \bibinfo{author}{S.~Ma},
  \bibinfo{author}{Z.~Huang}, \bibinfo{author}{A.~Karpathy},
  \bibinfo{author}{A.~Khosla}, \bibinfo{author}{M.~Bernstein}, et~al.,
  \bibinfo{title}{Imagenet large scale visual recognition challenge},
  \bibinfo{journal}{International Journal of Computer Vision}
  \bibinfo{volume}{115}~(\bibinfo{number}{3}) (\bibinfo{year}{2015})
  \bibinfo{pages}{211--252}.

\bibitem[{Xiao et~al.(2014)Xiao, Zhang, Yang, Peng, and Zhang}]{Xiao2014}
\bibinfo{author}{T.~Xiao}, \bibinfo{author}{J.~Zhang},
  \bibinfo{author}{K.~Yang}, \bibinfo{author}{Y.~Peng},
  \bibinfo{author}{Z.~Zhang}, \bibinfo{title}{{Error-Driven Incremental
  Learning in Deep Convolutional Neural Network for Large-Scale Image
  Classification}}, \bibinfo{journal}{MM '14 Proceedings of the ACM
  International Conference on Multime} \bibinfo{volume}{d}
  (\bibinfo{year}{2014}) \bibinfo{pages}{177--186}.

\bibitem[{Goodfellow et~al.(2013{\natexlab{a}})Goodfellow, Mirza, Xiao,
  Courville, and Bengio}]{Goodfellow2013}
\bibinfo{author}{I.~J. Goodfellow}, \bibinfo{author}{M.~Mirza},
  \bibinfo{author}{D.~Xiao}, \bibinfo{author}{A.~Courville},
  \bibinfo{author}{Y.~Bengio}, \bibinfo{title}{An empirical investigation of
  catastrophic forgetting in gradient-based neural networks},
  \bibinfo{journal}{arXiv preprint arXiv:1312.6211} .

\bibitem[{John~Walker(2014)}]{john2014big}
\bibinfo{author}{S.~John~Walker}, \bibinfo{title}{Big data: A revolution that
  will transform how we live, work, and think}, \bibinfo{year}{2014}.

\bibitem[{Fei-Fei et~al.(2006)Fei-Fei, Fergus, and Perona}]{fei2006one}
\bibinfo{author}{L.~Fei-Fei}, \bibinfo{author}{R.~Fergus},
  \bibinfo{author}{P.~Perona}, \bibinfo{title}{One-shot learning of object
  categories}, \bibinfo{journal}{IEEE transactions on pattern analysis and
  machine intelligence} \bibinfo{volume}{28}~(\bibinfo{number}{4})
  (\bibinfo{year}{2006}) \bibinfo{pages}{594--611}.

\bibitem[{Girshick(2015)}]{Girshick_2015}
\bibinfo{author}{R.~Girshick}, \bibinfo{title}{Fast R-{CNN}}, in:
  \bibinfo{booktitle}{2015 {IEEE} International Conference on Computer Vision
  ({ICCV})}, \bibinfo{publisher}{{IEEE}}, \bibinfo{year}{2015}.

\bibitem[{Shmelkov et~al.(2017)Shmelkov, Schmid, and Alahari}]{Shmelkov_2017}
\bibinfo{author}{K.~Shmelkov}, \bibinfo{author}{C.~Schmid},
  \bibinfo{author}{K.~Alahari}, \bibinfo{title}{Incremental Learning of Object
  Detectors without Catastrophic Forgetting}, in: \bibinfo{booktitle}{2017
  {IEEE} International Conference on Computer Vision ({ICCV})},
  \bibinfo{publisher}{{IEEE}}, \bibinfo{year}{2017}.

\bibitem[{Li and Hoiem(2017)}]{li2017learning}
\bibinfo{author}{Z.~Li}, \bibinfo{author}{D.~Hoiem}, \bibinfo{title}{Learning
  without forgetting}, \bibinfo{journal}{IEEE Transactions on Pattern Analysis
  and Machine Intelligence} .

\bibitem[{Aljundi et~al.(2016)Aljundi, Chakravarty, and
  Tuytelaars}]{aljundi2016expert}
\bibinfo{author}{R.~Aljundi}, \bibinfo{author}{P.~Chakravarty},
  \bibinfo{author}{T.~Tuytelaars}, \bibinfo{title}{Expert gate: Lifelong
  learning with a network of experts}, \bibinfo{journal}{CoRR, abs/1611.06194}
  \bibinfo{volume}{2}.

\bibitem[{Rusu et~al.(2016)Rusu, Rabinowitz, Desjardins, Soyer, Kirkpatrick,
  Kavukcuoglu, Pascanu, and Hadsell}]{rusu2016progressive}
\bibinfo{author}{A.~A. Rusu}, \bibinfo{author}{N.~C. Rabinowitz},
  \bibinfo{author}{G.~Desjardins}, \bibinfo{author}{H.~Soyer},
  \bibinfo{author}{J.~Kirkpatrick}, \bibinfo{author}{K.~Kavukcuoglu},
  \bibinfo{author}{R.~Pascanu}, \bibinfo{author}{R.~Hadsell},
  \bibinfo{title}{Progressive neural networks}, \bibinfo{journal}{arXiv
  preprint arXiv:1606.04671} .

\bibitem[{Rebuffi et~al.(2017)Rebuffi, Kolesnikov, and
  Lampert}]{rebuffi2017icarl}
\bibinfo{author}{S.-A. Rebuffi}, \bibinfo{author}{A.~Kolesnikov},
  \bibinfo{author}{C.~H. Lampert}, \bibinfo{title}{iCaRL: Incremental
  classifier and representation learning}, in: \bibinfo{booktitle}{Proc. CVPR},
  \bibinfo{year}{2017}.

\bibitem[{Sarwar et~al.(2017{\natexlab{a}})Sarwar, Panda, and Roy}]{Sarwar2017}
\bibinfo{author}{S.~S. Sarwar}, \bibinfo{author}{P.~Panda},
  \bibinfo{author}{K.~Roy}, \bibinfo{title}{{Gabor filter assisted energy
  efficient fast learning Convolutional Neural Networks}}, in:
  \bibinfo{booktitle}{2017 IEEE/ACM International Symposium on Low Power
  Electronics and Design (ISLPED)}, \bibinfo{publisher}{IEEE},
  \bibinfo{pages}{1--6}, \bibinfo{year}{2017}{\natexlab{a}}.

\bibitem[{Yosinski et~al.(2014)Yosinski, Clune, Bengio, and
  Lipson}]{yosinski2014transferable}
\bibinfo{author}{J.~Yosinski}, \bibinfo{author}{J.~Clune},
  \bibinfo{author}{Y.~Bengio}, \bibinfo{author}{H.~Lipson}, \bibinfo{title}{How
  transferable are features in deep neural networks?}, in:
  \bibinfo{booktitle}{Advances in neural information processing systems},
  \bibinfo{pages}{3320--3328}, \bibinfo{year}{2014}.

\bibitem[{Sarwar et~al.(2017{\natexlab{b}})Sarwar, Ankit, and
  Roy}]{sarwar2017incremental}
\bibinfo{author}{S.~S. Sarwar}, \bibinfo{author}{A.~Ankit},
  \bibinfo{author}{K.~Roy}, \bibinfo{title}{Incremental Learning in Deep
  Convolutional Neural Networks Using Partial Network Sharing},
  \bibinfo{journal}{arXiv preprint arXiv:1712.02719} .

\bibitem[{Panda and Roy(2017)}]{panda2017semantic}
\bibinfo{author}{P.~Panda}, \bibinfo{author}{K.~Roy}, \bibinfo{title}{Semantic
  driven hierarchical learning for energy-efficient image classification}, in:
  \bibinfo{booktitle}{2017 Design, Automation \& Test in Europe Conference \&
  Exhibition (DATE)}, \bibinfo{organization}{IEEE},
  \bibinfo{pages}{1582--1587}, \bibinfo{year}{2017}.

\bibitem[{Panda et~al.(2017)Panda, Ankit, Wijesinghe, and
  Roy}]{panda2017falcon}
\bibinfo{author}{P.~Panda}, \bibinfo{author}{A.~Ankit},
  \bibinfo{author}{P.~Wijesinghe}, \bibinfo{author}{K.~Roy},
  \bibinfo{title}{FALCON: Feature driven selective classification for
  energy-efficient image recognition}, \bibinfo{journal}{IEEE Transactions on
  Computer-Aided Design of Integrated Circuits and Systems}
  \bibinfo{volume}{36}~(\bibinfo{number}{12}).

\bibitem[{Yan et~al.(2015)Yan, Zhang, Piramuthu, Jagadeesh, DeCoste, Di, and
  Yu}]{Yan_2015}
\bibinfo{author}{Z.~Yan}, \bibinfo{author}{H.~Zhang},
  \bibinfo{author}{R.~Piramuthu}, \bibinfo{author}{V.~Jagadeesh},
  \bibinfo{author}{D.~DeCoste}, \bibinfo{author}{W.~Di},
  \bibinfo{author}{Y.~Yu}, \bibinfo{title}{{HD}-{CNN}: Hierarchical Deep
  Convolutional Neural Networks for Large Scale Visual Recognition}, in:
  \bibinfo{booktitle}{2015 {IEEE} International Conference on Computer Vision
  ({ICCV})}, \bibinfo{publisher}{{IEEE}}, \bibinfo{year}{2015}.

\bibitem[{Srivastava and Salakhutdinov(2013)}]{srivastava2013discriminative}
\bibinfo{author}{N.~Srivastava}, \bibinfo{author}{R.~R. Salakhutdinov},
  \bibinfo{title}{Discriminative transfer learning with tree-based priors}, in:
  \bibinfo{booktitle}{Advances in Neural Information Processing Systems},
  \bibinfo{pages}{2094--2102}, \bibinfo{year}{2013}.

\bibitem[{Kontschieder et~al.(2015)Kontschieder, Fiterau, Criminisi, and
  Bulo}]{kontschieder2015deep}
\bibinfo{author}{P.~Kontschieder}, \bibinfo{author}{M.~Fiterau},
  \bibinfo{author}{A.~Criminisi}, \bibinfo{author}{S.~R. Bulo},
  \bibinfo{title}{Deep neural decision forests}, in:
  \bibinfo{booktitle}{Computer Vision (ICCV), 2015 IEEE International
  Conference on}, \bibinfo{organization}{IEEE}, \bibinfo{pages}{1467--1475},
  \bibinfo{year}{2015}.

\bibitem[{Simonyan and Zisserman(2014)}]{simonyan2014very}
\bibinfo{author}{K.~Simonyan}, \bibinfo{author}{A.~Zisserman},
  \bibinfo{title}{Very deep convolutional networks for large-scale image
  recognition}, \bibinfo{journal}{arXiv preprint arXiv:1409.1556} .

\bibitem[{Vedaldi and Lenc(2015)}]{Vedaldi_2015}
\bibinfo{author}{A.~Vedaldi}, \bibinfo{author}{K.~Lenc},
  \bibinfo{title}{Matconvnet: Convolutional neural networks for matlab}, in:
  \bibinfo{booktitle}{Proceedings of the 23rd ACM international conference on
  Multimedia}, \bibinfo{organization}{ACM}, \bibinfo{pages}{689--692},
  \bibinfo{year}{2015}.

\bibitem[{MATLAB(2017)}]{MATLAB_2017}
\bibinfo{author}{MATLAB}, \bibinfo{title}{version 9.2.0 (R2017a)},
  \bibinfo{publisher}{The MathWorks Inc.}, \bibinfo{address}{Natick,
  Massachusetts}, \bibinfo{year}{2017}.

\bibitem[{Goodfellow et~al.(2013{\natexlab{b}})Goodfellow, Warde-Farley, Mirza,
  Courville, and Bengio}]{goodfellow2013maxout}
\bibinfo{author}{I.~J. Goodfellow}, \bibinfo{author}{D.~Warde-Farley},
  \bibinfo{author}{M.~Mirza}, \bibinfo{author}{A.~Courville},
  \bibinfo{author}{Y.~Bengio}, \bibinfo{title}{Maxout networks}, in:
  \bibinfo{booktitle}{International Conference on Machine Learning},
  \bibinfo{year}{2013}{\natexlab{b}}.

\bibitem[{Srivastava et~al.(2014)Srivastava, Hinton, Krizhevsky, Sutskever, and
  Salakhutdinov}]{srivastava2014dropout}
\bibinfo{author}{N.~Srivastava}, \bibinfo{author}{G.~E. Hinton},
  \bibinfo{author}{A.~Krizhevsky}, \bibinfo{author}{I.~Sutskever},
  \bibinfo{author}{R.~Salakhutdinov}, \bibinfo{title}{Dropout: a simple way to
  prevent neural networks from overfitting.}, \bibinfo{journal}{Journal of
  machine learning research} \bibinfo{volume}{15}~(\bibinfo{number}{1})
  (\bibinfo{year}{2014}) \bibinfo{pages}{1929--1958}.

\bibitem[{Ioffe and Szegedy(2015)}]{ioffe2015batch}
\bibinfo{author}{S.~Ioffe}, \bibinfo{author}{C.~Szegedy}, \bibinfo{title}{Batch
  normalization: Accelerating deep network training by reducing internal
  covariate shift}, in: \bibinfo{booktitle}{International Conference on Machine
  Learning}, \bibinfo{pages}{448--456}, \bibinfo{year}{2015}.

\bibitem[{Hertel et~al.(2015)Hertel, Barth, Kaster, and
  Martinetz}]{Hertel_2015}
\bibinfo{author}{L.~Hertel}, \bibinfo{author}{E.~Barth},
  \bibinfo{author}{T.~Kaster}, \bibinfo{author}{T.~Martinetz},
  \bibinfo{title}{Deep convolutional neural networks as generic feature
  extractors}, in: \bibinfo{booktitle}{2015 International Joint Conference on
  Neural Networks ({IJCNN})}, \bibinfo{publisher}{{IEEE}},
  \bibinfo{year}{2015}.

\bibitem[{He et~al.(2016)He, Zhang, Ren, and Sun}]{he2016deep}
\bibinfo{author}{K.~He}, \bibinfo{author}{X.~Zhang}, \bibinfo{author}{S.~Ren},
  \bibinfo{author}{J.~Sun}, \bibinfo{title}{Deep residual learning for image
  recognition}, in: \bibinfo{booktitle}{Proceedings of the IEEE conference on
  computer vision and pattern recognition}, \bibinfo{pages}{770--778},
  \bibinfo{year}{2016}.

\end{thebibliography}

\end{document}